\documentclass[times,twocolumn,final]{elsarticle}

\usepackage{medima}
\usepackage{framed,multirow}

\usepackage{amssymb}
\usepackage{latexsym}
\usepackage{amsmath}

\usepackage{url}
\usepackage[table]{xcolor}
\usepackage{xcolor}

\usepackage{hyperref}

\definecolor{newcolor}{rgb}{.8,.349,.1}

\journal{Preprint}

\begin{document}

\verso{Moi Hoon Yap \textit{et~al.}}

\begin{frontmatter}

\title{Deep Learning in Diabetic Foot Ulcers Detection: A Comprehensive Evaluation}

\author[1]{Moi Hoon \snm{Yap}\corref{cor1}}
\cortext[cor1]{Corresponding author: 
	Tel.: +44 161 247 1503;  
}
\ead{M.Yap@mmu.ac.uk}
\author[2]{Ryo \snm{Hachiuma}\fnref{fn1}}
\fntext[fn1]{Authors with equal contribution.}
\author[3]{Azadeh \snm{Alavi}\fnref{fn1}}
\author[4]{Raphael \snm{Br\"ungel}\fnref{fn1}}
\author[1]{Bill \snm{Cassidy}}
\author[5]{Manu \snm{Goyal}\fnref{fn1}}
\author[6]{Hongtao \snm{Zhu}\fnref{fn1}}
\author[4]{Johannes \snm{R\"uckert}}
\author[3]{Moshe \snm{Olshansky}}
\author[6]{Xiao \snm{Huang}}
\author[2]{Hideo \snm{Saito}}
\author[5]{Saeed \snm{Hassanpour}}
\author[4,7]{Christoph M. \snm{Friedrich}}
\author[3]{David \snm{Ascher}}
\author[6]{Anping \snm{Song}}
\author[8]{Hiroki \snm{Kajita}}
\author[1]{David \snm{Gillespie}}
\author[1]{Neil D. \snm{Reeves}}
\author[9]{Joseph \snm{Pappachan}}
\author[10]{Claire \snm{O'Shea}}
\author[11]{Eibe \snm{Frank}}

\address[1]{Manchester Metropolitan University, John Dalton Building, Chester Street, Manchester M1 5GD, UK}
\address[2]{Keio University, Yokohama, Kanagawa, Japan}
\address[3]{Baker Heart and Diabetes Institute, 20 Commercial Road, Melbourne, VIC 3000, Australia}
\address[4]{Department of Computer Science, University of Applied Sciences and Arts
	Dortmund (FH Dortmund), Emil-Figge-Str. 42, 44227 Dortmund, Germany}
\address[5]{Department of Biomedical Data Science, Dartmouth College, Hanover, NH, USA}
\address[6]{Shanghai University, Shanghai 200444, China}
\address[7]{Institute for Medical Informatics, Biometry and Epidemiology (IMIBE), University Hospital Essen, Hufelandstr. 55, 45122 Essen, Germany}
\address[8]{Keio University School of Medicine, Shinanomachi, Tokyo, Japan}
\address[9]{Lancashire Teaching Hospitals, Chorley, UK}
\address[10]{Waikato Diabetes Health Board, Hamilton, New Zealand}
\address[11]{Department of Computer Science, University of Waikato, Hamilton, New Zealand}


\begin{abstract}
There has been a substantial amount of research involving computer methods and technology for the detection and recognition of diabetic foot ulcers (DFUs), but there is a lack of systematic comparisons of state-of-the-art deep learning object detection frameworks applied to this problem. DFUC2020 provided participants with a comprehensive dataset consisting of 2,000 images for training and 2,000 images for testing. This paper summarises the results of DFUC2020 by comparing the deep learning-based algorithms proposed by the winning teams: Faster R-CNN, three variants of Faster R-CNN and an ensemble method; YOLOv3; YOLOv5; EfficientDet; and a new Cascade Attention Network. For each deep learning method, we provide a detailed description of model architecture, parameter settings for training and additional stages including pre-processing, data augmentation and post-processing. We provide a comprehensive evaluation for each method. All the methods required a data augmentation stage to increase the number of images available for training and a post-processing stage to remove false positives. The best performance was obtained from Deformable Convolution, a variant of Faster R-CNN, with a mean average precision (mAP) of 0.6940 and an F1-Score of 0.7434. Finally, we demonstrate that the ensemble method based on different deep learning methods can enhanced the F1-Score but not the mAP.

\end{abstract}


\end{frontmatter}


\section{Introduction} 
\label{sec1}
According to the International Diabetes Federation \cite{IDF}, in 2019 there were approximately 463 million adults with diabetes worldwide. This number is expected to grow to 700 million by 2045. A person with diabetes has a 34\% lifetime risk of developing a diabetic foot ulcer (DFU). In other words, 1 in every 3 people with diabetes will develop a DFU in their lifetime \cite{armstrong2017diabetic}. Infection of a DFU frequently leads to limb amputation, causing significant morbidity, psychological distress and reduced quality of life and life expectancy. This research is the first step of a future diabetic foot care project. Periodic monitoring of foot ulcers is important to assess the progress of ulcer healing, which is currently performed manually by clinicians. Many foot clinics take photographs of ulcers during initial evaluation and subsequent reviews for comparison of various stages of ulcer progression to boost the visual memory of clinicians. The current research aims to develop artificial intelligence-based deep learning algorithms for detection of ulcers without direct clinical intervention. This is especially important in the current COVID-19 climate, where social distancing is of paramount importance. Technologies developed to enhance ulcer diagnostics and care plans have the potential to revolutionise diabetic foot care. 

Detection tasks can be challenging when taking into account the numerous environmental elements in real-world settings. Examples of some observations include:

\begin{itemize}
	\item Newly acquired and subtle early stages of ulceration can easily be missed by care personnel during visual assessment of priorly acquired conditions due to the short time designated for standard treatment
	\item Low-quality images with bad focus, motion blur, occlusion, poor lighting, and backlight are common in wound documentation due to limited available time for treatment and documentation, even when performed by trained personnel
	\item Malformed toenails, deep rhagades, folded amputation scars, and fresh epithelialization are examples for false positive detections that require manual correction, which can be time consuming when documenting DFU
	\item Very small and very large and curved ulcers are problematic for certain detectors, but are common in typical wound care documentation
\end{itemize}

It is essential to develop a technological solution capable of transforming current screening practices that have the potential to significantly reduce clinical time burdens. 

With the emerging growth of deep learning, automated analysis of DFU has become possible. However, deep learning requires large-scale datasets to achieve results comparable with those of human experts. Currently, medical imaging researchers are working in isolation and the majority of their research is not reproducible. To bridge the gap and to motivate data sharing amongst researchers and clinicians, Yap et al. \cite{yap2020dfuc, yap2020dfuc2021} proposed the diabetic foot ulcer challenges. This paper presents an overview of the state-of-the-art computer methods in DFU detection, provides an overview of the publicly available datasets, presents a comprehensive evaluation of the popular object detection frameworks on DFU detection, proposes an ensemble method and Cascade Attention DetNet for DFU detection, and conducts a comprehensive evaluation of the deep learning algorithms trained on the DFUC2020 dataset.

\section{Related Work} 
\label{sec2}

The growing number of reported cases of diabetes has resulted in a corresponding growth in research interest in DFU. Early attempts in training deep learning models in this domain have shown promising results. Previous research \cite{goyal2018dfunet,goyal2017fully,goyal2018robust} trained models capable of classification, localisation and segmentation. These models reported high levels of mean average precision (mAP), sensitivity and specificity in experimental settings. The existing method on localisation was trained using Faster R-CNN with Inception v2 and two-tier transfer learning from the Microsoft Common Objects in Context (MS COCO) dataset. However, despite the high scoring performance measures, these models were trained and evaluated on small datasets ($<$2000), therefore the results cannot be regarded as conclusive evidence of their efficacy in real-world settings.

Brown et al. \cite{brown2017myfootcare} created the MyFootCare mobile app which was designed to encourage patient self-monitoring using diaries, goals and notifications. The app stores a log of patient foot images and is capable of semi-automated segmentation. This novel solution to maintaining foot records utilises a method of automatic photograph capture where the phone is placed on the floor and the patient is guided using voice feedback. However, this particular function of the system was not tested during the actual experiment, so it is not known how well it performed in real-world settings.

Wang et al. \cite{wang2014smartphone,wang2016area} devised a method of consistent DFU image capture using a box with a glass surface containing mirrors which reflect the image back to a camera or mobile device. Cascaded two-stage support vector classification was used to ascertain the DFU region, followed by a two-stage super-pixel classification technique used for segmentation and feature extraction. Despite being highly novel, this method exhibited a number of limitations, such as risk of infection due to physical contact between wound and capture box. The design of the capture box also limited monitoring to DFU that are present on the plantar surface of the foot. The sample size was also statistically insignificant, with only 35 images from real patients and 30 images of wound moulds.

\section{Datasets} 
\label{sec3}
The DFU datasets provided by The Manchester Metropolitan University and Lancashire Teaching Hospitals NHS Trust \cite{goyal2018dfunet, goyal2020recognition, cassidy2020dfuc2020} are digital DFU image datasets with expert annotations. The aim of the publication of this data is to encourage more researchers to work in this domain and to conduct reproducible experiments. There are three types of datasets made publicly available for researchers. The first dataset consists of foot skin patches for wound classification \cite{goyal2018dfunet}; the second dataset contains regions of interests for infection and ischaemia classification \cite{goyal2020recognition}; and the third is the most recently published dataset for DFU detection \cite{cassidy2020dfuc2020}. The third dataset is the largest dataset to date, and increased usage of this data is the driving force for the organisers of the DFU challenges. The researchers involved in organising the yearly DFU challenges \cite{yap2020dfuc, yap2020dfuc2021}, in conjunction with the MICCAI conferences, aim to attract wider participation to improve the diagnosis/monitoring of foot ulcers and to raise awareness of diabetes and DFU. There are numerous aspects to take into account in the development of accurate detection algorithms. As is the case with other medical imaging research fields, increasing the number of images is only one of them. The Diabetic Foot Ulcers Grand Challenge (DFUC2020) dataset consist of 2,000 training images, 200 validation images and 2,000 testing images \cite{cassidy2020dfuc2020, goyal2018robust}. The data consists of 2,496 ulcers in the training set and 2,097 ulcers in the testing set. In an attempt to promote model robustness, some of the images in the testing set do not exhibit DFUs. The details of the dataset are described in \cite{cassidy2020dfuc2020}. To improve the performance of the deep learning methods and to reduce computational costs, all images were resized to 640 $\times$ 480 pixels. 

\begin{figure}
	\centering
	\includegraphics[scale=0.3]{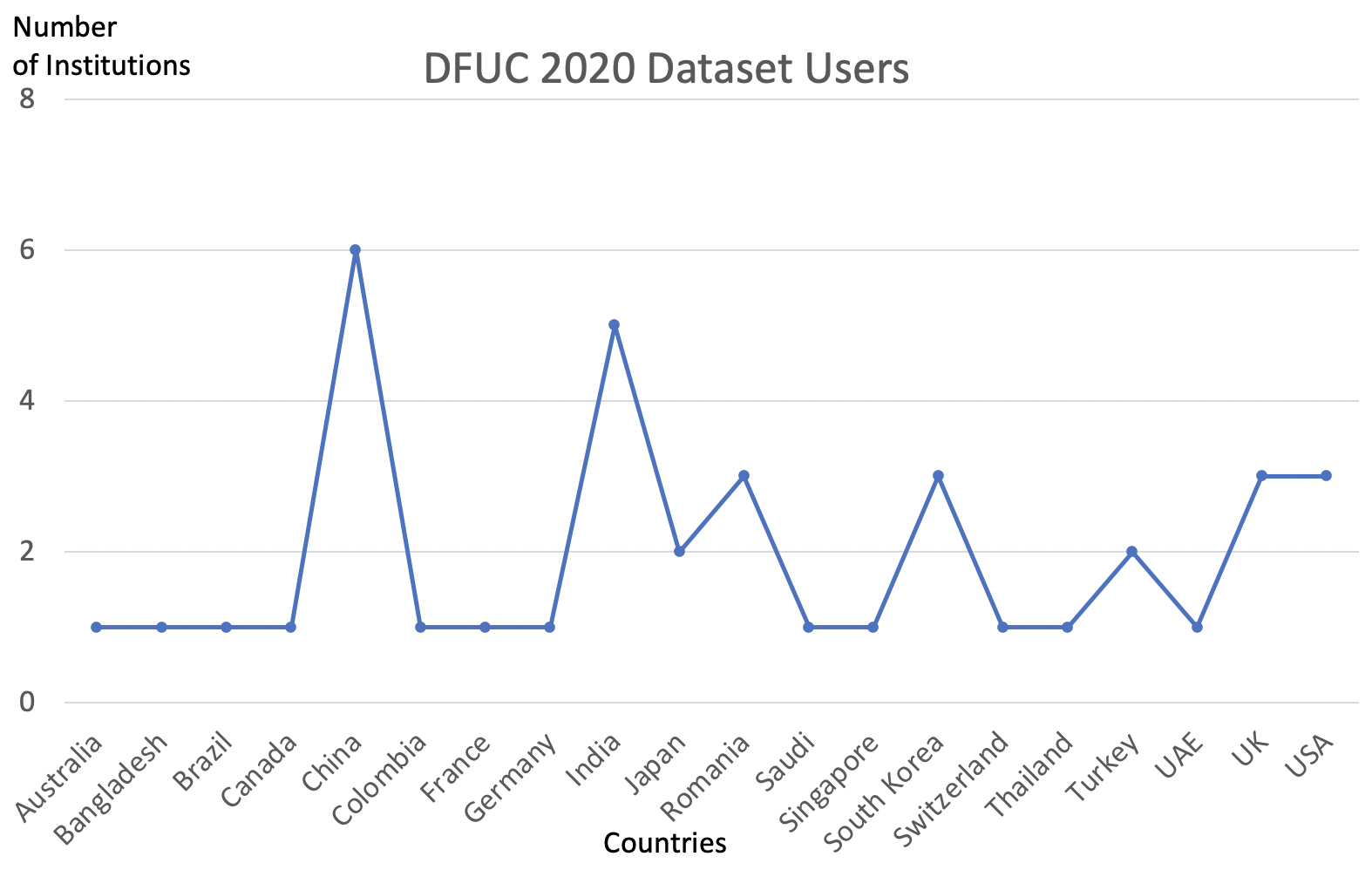}
	\caption{Summary of DFUC2020 participants across the world.}
	\label{fig:users}
\end{figure}

Since the release of the DFUC2020 training dataset on the 27th April 2020, we received requests from 39 international institutions, as shown in Fig. \ref{fig:users}. There are a total of 31 submissions to the challenge from 11 teams. We report the top scores from each team and discuss their methods according to the object detection approaches they implemented. 

\section{DFU Detection Methods} 
This section presents a comprehensive description of the DFU detection methods used, grouped according to the popular deep learning object detection algorithms they apply, i.e.  Faster R-CNN, YOLOv3, YOLOv5 and EfficientDet. We also include descriptions of an ensemble method and a new Cascade Attention DetNet (CA-DetNet).

\subsection{Faster R-CNN} 
Faster R-CNN \cite{ren2015faster} is one of the two-stage object detection models, which generates a sparse set of candidate object locations using a Region Pooling Network (RPN) based on shared feature maps, which then classifies each candidate proposal as the foreground or background class. After extracting shared feature maps with a CNN, the first stage RPN takes shared feature maps as an input and generates a set of bounding box candidate object locations, each with an "objectness" score. The size of each anchor is configured using hyperparameters. Then, the proposals are used in the region of interest pooling layer (RoI pooling) to generate subfeature maps. The subfeature maps are converted to 4,096 dimensional vectors and fed forward into fully connected layers. These layers are then used as a regression network to predict bounding box offsets, with a classification network used to predict the class label of each bounding box proposal.

The RoI pooling layer quantizes a floating-number RoI to the discrete granularity of the feature map. This quantization introduces misalignments between the RoI and the extracted features. Therefore, the model evaluated in this paper employs a RoIAlign layer, which is introduced in Mask R-CNN \cite{MaskRCNN}, instead of the RoI pooling layer. This removes the harsh quantization of the RoI pooling layer, properly aligning the extracted features with the input.

Additionally, the Feature Pyramid Network (FPN) \cite{aa_lin2017feature} is employed as the backbone of the network. FPN uses a top-down architecture with lateral connections to build an in-network feature pyramid from a single-scale input. Faster R-CNN with an FPN backbone extracts RoI features from different levels of the feature pyramid according to their scale, with the remainder of the approach being similar to ResNet. Using a ResNet-FPN backbone for feature extraction with Mask R-CNN gives excellent gains in both accuracy and speed. Specifically, we employ ResNeXt101 \cite{Aggregated2017Xie} with the FPN feature extraction backbone to extract the features. 

\subsubsection{Data Augmentation}
In this challenge, the images in the dataset were captured from different viewpoint angles, cameras with different focal lengths and varying levels of blur. Also, the training dataset contains only $2,000$ images, which could be considered small for training deep learning models. Therefore, we employ various data augmentation techniques for robust prediction. Specifically, we employ the following augmentations:

\begin{itemize}
	\item HSV and RGB: As the lighting conditions vary between dataset images, we apply random RGB and HSV shift to the images. Especially, we randomly add/subtract from $0$ to $10$ RGB values and $0$ to $20$ HSV values in the images.
	\item Blurring: As the dataset contains images captured from different focal lengths, some images are blurred and contain camera noise. Therefore, we apply Gaussian and median blur filters with the filter size set to $3$. The filters are applied with the probability of $0.1$.
	\item Affine transformation: As the images are captured from different camera angles, we apply random affine transformations. Specifically, we apply random shift, scaling ($0.1$) and rotation ($90$ degrees). 
	\item Brightness: As the images are captured in various environments, we employ brightness and contrast data augmentation. More specifically, we randomly change the brightness and contrast in a scale from $0.1$ to $0.3$, with probability set to $0.2$. 
\end{itemize}

\subsubsection{Model training and implementation}
In this paper, we fine-tune a model pretrained on MS-COCO \cite{lin2014microsoft}. We employ Stochastic Gradient Descent Optimizer with a momentum of $0.9$ and weight decay set to $0.0001$. During training, we employ a warm up learning rate scheduling strategy, using lower learning rates in the early stages of training to overcome optimization difficulties. More specifically, we linearly increase the learning rate to $0.01$ in the first $500$ iterations, then multiply by $0.1$ at epochs $6, 12$ and $30$. We implemented the methods based on the mmdetection repository \footnote{\url{https://github.com/open-mmlab/mmdetection}}. 

\subsubsection{Variants of Faster R-CNN}
Several papers have proposed variants of Faster R-CNN. In this paper, we implement Faster R-CNN, three variants of Faster R-CNN and ensemble the results. The three variants of Faster R-CNN are as follows:

\begin{itemize}
	\item Cascade R-CNN \cite{cai2019cascade}: this variant implements a different  architecture for the ROI head (the module that predicts the bounding boxes and the category label). Cascade R-CNN builds up a cascade head based on Faster R-CNN \cite{ren2015faster} to refine detection progressively. Since the proposal boxes are refined by multiple box regression heads, Cascade R-CNN is optimal for more precise localization of objects. 
	
	\item Deformable Convolution \cite{Zhu2019Deformable}: in this variant, the basic architecture of the network is the same as Faster R-CNN. However, we replace the convolution layer with a deformable convolution layer \cite{zhu2018deformable} at the second, third and fourth ResNeXt blocks of the feature extractor. The deformable convolution adds 2D offsets to the regular grid sampling locations in the standard convolution, enabling free-form deformation of the sampling grid. The offsets are learned from the feature maps, via additional convolutional layers. Thus, the deformation is conditioned on the input features in a local, dense and adaptive manner. 
	
	\item Prime Sample Attention \cite{Cao2020PISA} (PISA): PISA is motivated by two considerations: samples should not be treated as independent and equally important, and the classification and localization are correlated. Thus, it employs a ranking strategy that places the positive samples with highest IoUs around each object, and the negative samples with highest scores in each cluster at the top of the ranked list. This directs the focus of the training process via a simple re-weighting scheme. It also employs a classification-aware regression loss to jointly optimize the classification and regression branches.
\end{itemize}

\subsubsection{Post-processing}
At test time, we employ a test-time augmentation scheme: we augment the test image by applying two resolutions, and we also flip the image. As a result, we augment a single image to four images and merge the predictions obtained for the four images. We employ soft NMS (non maximum suppression) \cite{bodla2017soft} with a confidence threshold of $0.5$ as the post-processing of predicted bounding boxes. 

\subsubsection{Ensemble method}
Combining predictions from different models can improve generalization and usually yields more accurate results compared to a single model. During the post-processing stage for Faster R-CNNs, we employ soft NMS \cite{bodla2017soft} to select the predicted bounding boxes for each method. Such methods work well on a single model, but they only select the boxes and cannot produce averaged localization of predictions combined from various models effectively. Therefore, after predicting the bounding boxes for each method, we ensemble these predicted bounding boxes using Weighted Boxes Fusion \cite{solovyev2019weighted}. Unlike NMS-based methods that simply exclude part of the predicted bounding boxes, the Weighted Boxes Fusion algorithm uses the confidence scores of all proposed bounding boxes to form the average boxes. The reader is referred to \cite{solovyev2019weighted} for further details of the algorithm. We ensemble four models (pure Faster R-CNN, Cascade R-CNN, Faster R-CNN with Deformable Convolution and Faster R-CNN with Prime Sample Attention model). We set equal weights when fusing the predicted bounding boxes of each model.

\subsection{YOLO} 
You-Only-Look-Once (YOLO) \cite{yolov1} is a unified, real-time object detection algorithm that reformulates the object detection task to a single regression problem. YOLO employs a single neural network architecture to predict bounding boxes and class probabilities directly from full images. Hence, when compared to Faster R-CNN \cite{ren2015faster}, YOLO provides faster detection.

Over time, improvements of YOLO were implemented and released as distinct and independent software packages by the originators \cite{yolov1, yolov2, aa_redmon2018yolov3}. As a result of increased publicity and popularity, a model zoo containing further YOLO adaptations emerged. Subsequently, further maintainers continued to improve the DarkNet\footnote{DarkNet GitHub repository: \url{https://github.com/pjreddie/darknet} (accessed 2020-08-29)}-based versions, and \cite{yolov4} created ports for other machine learning libraries such as PyTorch\footnote{PyTorch website: \url{https://pytorch.org/} (accessed 2020-08-29)} \cite{pytorch}.

In this paper, two approaches are selected for DFU detection using the DFUC2020 dataset: YOLOv3 and YOLOv5. We discuss the networks and present descriptions of our implementation in the following subsections.

\subsubsection{YOLOv3} 
YOLOv3 \cite{aa_redmon2018yolov3} was developed as an improved version of YOLOv2 \cite{yolov2}. It employs multi-scale schema, predicting bounding boxes on different scales. This allows YOLOv3 to be more effective for detecting smaller targets when compared to YOLOv2.

\begin{figure}[htb!]
	\centering
	\includegraphics[scale=0.35]{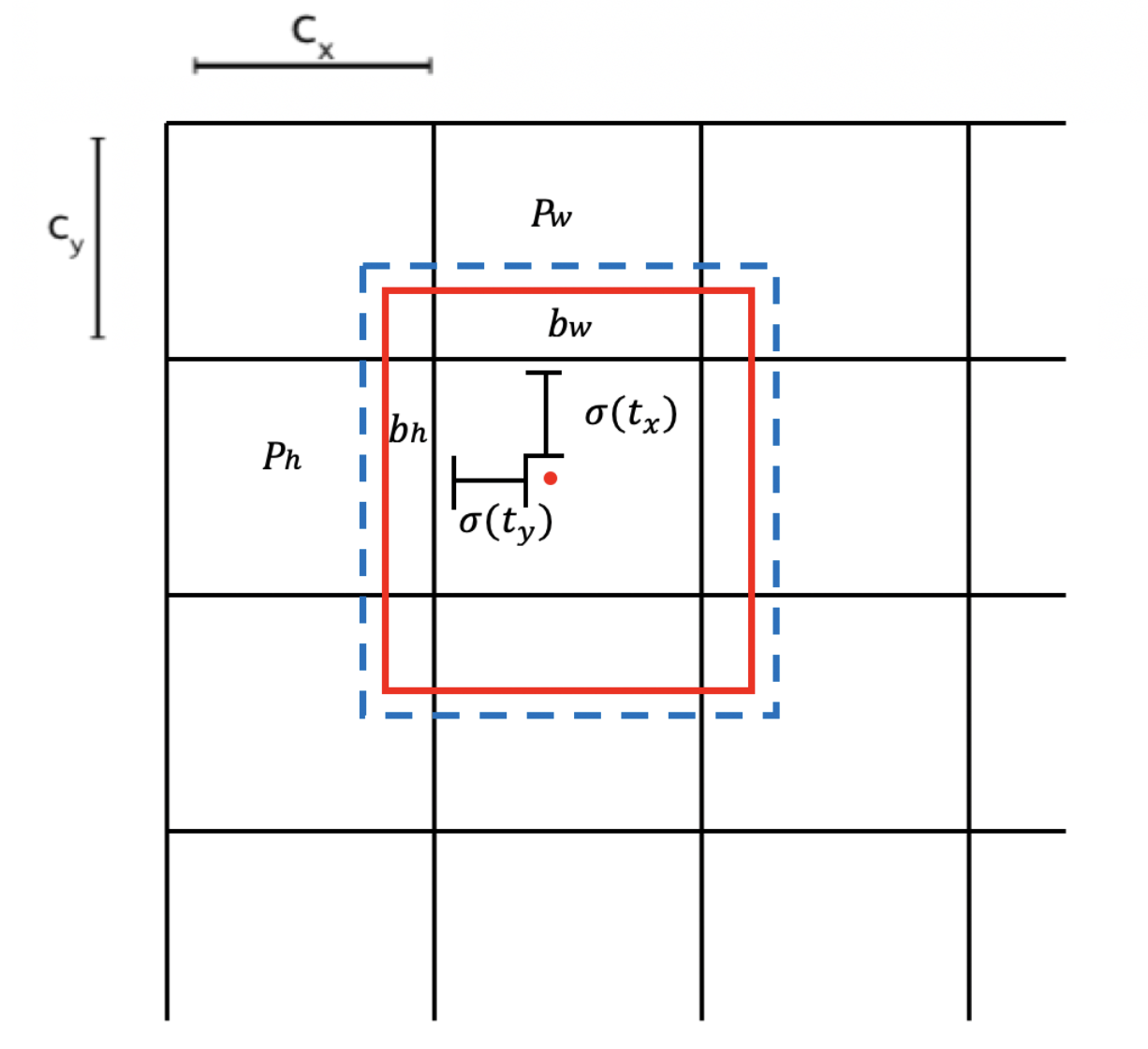}
	\caption{Illustration of bounding boxes, dimension priors and location prediction.}
	\label{fig:yolo3_bb}
\end{figure}

YOLOv3 uses dimension clusters as anchor boxes in order to predict bounding boxes around the desired objects in given images. Logistic regression is used to predict the objectness score for a given bounding box. Specifically, as illustrated in Fig. \ref{fig:yolo3_bb}, the algorithm predicts the four coordinates of the bounding box \((t_{x},t_{y},t_{h},t_{w})\) as in Equation \ref{eqn:yolov3}.

\begin{align}
	\label{eqn:yolov3}
	\begin{split}
		b_{x} = \sigma (t_{x})+ c_{x} \\
		b_{y} = \sigma (t_{y})+ c_{y} \\
		b_{h}= p_{w}e^{t_{w}} \\
		b_{w}= p_{h}e^{t_{h}}  
	\end{split}
\end{align} where \((c_{x},y_{y})\) are offsets from the top left corner of the image, and \((p_{w},p_{h})\) are bounding box prior height and weight. The k-means clustering algorithm is used to determine bounding box priors, while the sum of squared errors is used for training the network. Let \({\hat{t_{\ast}}}\) be the ground truth for some coordinate prediction, and \(t_{\ast}\) be the network prediction during training. Then, the gradient is \({\hat{t_{\ast}}} - t_{\ast}\), which can be easily computed by inverting equation \ref{eqn:yolov3}.

\subsubsection*{Model Pipeline}
The backbone of YOLOv3 is a hybrid model called Darknet-53 (as shown in Table \ref{table:DarkNet-53}), which is used for feature extraction. As the name indicates, DarkNet-53 is made of 53 convolutional layers that also take advantage of shortcut connections.

\begin{table}[htb!]
	\centering
	\small\addtolength{\tabcolsep}{3pt}
	\renewcommand{\arraystretch}{1.5}
	\caption{The architecture of DarkNet-53 used in YOLOv3.}
	\label{table:DarkNet-53}
	\scalebox{0.75}{
		\begin{tabular}{cccc}
			\hline
			Type &  &Filters  & Size            \\ \hline\hline
			Convolutional & &32&3$\times$3\\
			Convolutional & &64&3$\times$3/2\\
			\rowcolor{lightgray}
			Convolutional & &32&1$\times$1\\
			\rowcolor{lightgray}
			Convolutional &1$\times$ &64&3$\times$3\\
			\rowcolor{lightgray}
			Residual & & & \\
			Convolutional & &128&3$\times$3/2\\
			\rowcolor{lightgray}
			Convolutional & &64&1$\times$1\\
			\rowcolor{lightgray}
			Convolutional &2$\times$ &128&3$\times$3\\
			\rowcolor{lightgray}
			Residual & & & \\
			Convolutional & &256&3$\times$3/2\\
			\rowcolor{lightgray}
			Convolutional & &128&1$\times$1\\
			\rowcolor{lightgray}
			Convolutional &8$\times$ &256&3$\times$3\\
			\rowcolor{lightgray}
			Residual & & & \\
			Convolutional & &512&3$\times$3/2\\
			\rowcolor{lightgray}
			Convolutional & &256&1$\times$1\\
			\rowcolor{lightgray}
			Convolutional &8x &512&3$\times$3\\
			\rowcolor{lightgray}
			Residual & & & \\
			Convolutional & &1024&3$\times$3/2\\
			\rowcolor{lightgray}
			Convolutional & &512&1$\times$1\\
			\rowcolor{lightgray}
			Convolutional &4$\times$ &1024&3$\times$3\\
			\rowcolor{lightgray}
			Residual & & & \\
			\hline 
			Avgpool Connected Softmax &&Global 1000& \\
			\hline
			\hline
	\end{tabular}}
\end{table}

As the detection algorithm is required to detect only one type of object, the complexity of the problem is reduced from multi-class detection to single object detection. Hence, for the purpose of detecting diabetic foot ulcers, we have employed a simplified version of YOLOv3.

\subsubsection*{Training}
We employ transfer learning by using the pre-trained DarkNet weights which are provided by \cite{aa_redmon2018yolov3}. Then, we train our detector in 2 steps, using the following settings: Adam optimizer with learning rate 1e-3, number of epochs=100, batch size=32 and using 20\% of the data for validation.

First, we start by freezing the top DarkNet-53 layers and train the algorithm with the above settings. Then, we retrain the entire network to improve performance. Similar to the original YOLOv3, our trained network extracts features from 3 different pre-defined scales, which is a similar concept to feature pyramid networks \cite{aa_lin2017feature}. We then use the trained network for detecting diabetic foot ulcers in blind test images.

\subsubsection*{Post-processing}
\begin{figure}[!ht]
	\centering
	\includegraphics[scale=0.5]{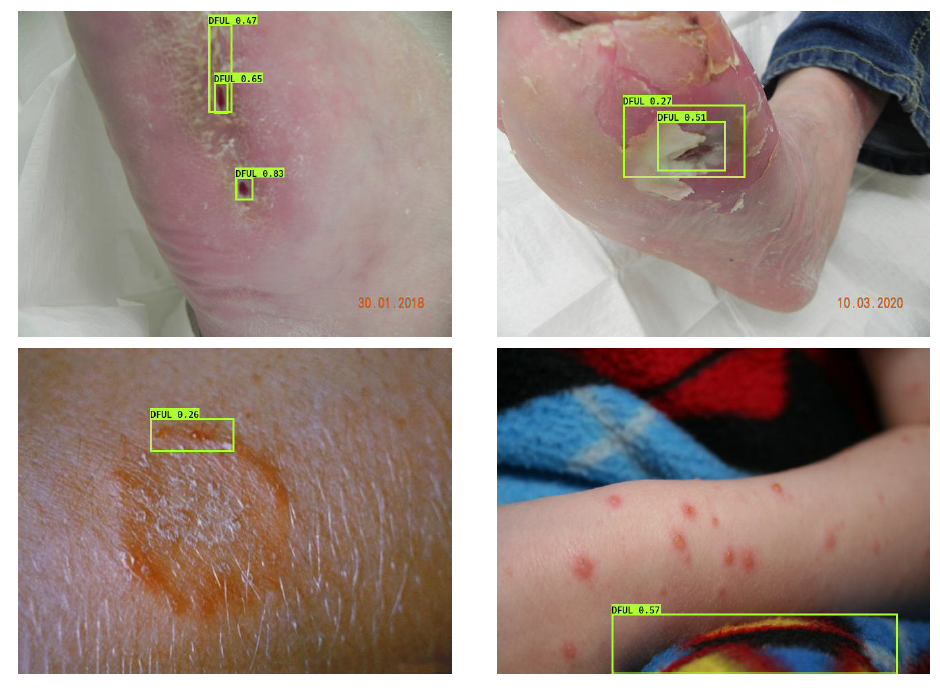}
	\caption{Illustration of two types of false positives: (top row) false positives from double detections; and (bottom row) single detection false positives.}
	\label{fig:yolo3_pp}
\end{figure}

As observed from Fig. \ref{fig:yolo3_pp}, in rare cases, the resulting algorithm may produce double detections or false positives. To reduce such examples, we include a post-processing stage.

Our post-processing steps consist of two stages. First, we identify double detections by flagging the detected bounding boxes with more that 80\% overlap. Among the overlapping detected boxes we only keep the box with the highest confidence result. Finally, we further post-process the results by removing any detection with a confidence score $<$ 0.3, with the aim of reducing the rate of false positive detections.

\subsubsection{YOLOv5} 

YOLOv5 was first published on GitHub\footnote{YOLOv5 GitHub repository: \url{https://github.com/ultralytics/yolov5/releases/tag/v1.0} (accessed 2021-04-28)} in May 2020 in v1.0 \cite{jocher2020a}. The maintainer is already well known for a YOLOv3 \cite{aa_redmon2018yolov3} port for PyTorch\footnote{Ultralytics' YOLOv3 GitHub repository: \url{https://github.com/ultralytics/yolov3} (accessed 2020-08-29)} \cite{jocher2020b}. The maintainer named the network YOLOv5 to avoid naming conflicts due to the prior release of YOLOv4 \cite{yolov4}. However, YOLOv5 is not to be confused with a descendent of the original DarkNet-based\footnote{YOLOv4 GitHub repository: \url{https://github.com/AlexeyAB/darknet} (accessed 2020-08-29)} YOLO-series. A scientific paper reporting on the improvements in YOLOv5 has not yet been published, but is currently pending\footnote{YOLOv5 question on scientific paper: \url{https://github.com/ultralytics/yolov5/issues/2847} (accessed 2021-04-28)}. YOLOv5 is currently under active development, with the latest version being v5.0 \cite{jocher2021} at the time of writing.

New features and improvements in YOLOv5 are mainly focused on the incorporation of the state-of-the-art for deep learning networks, such as activation functions and data augmentation. These were partly adopted from YOLOv4\footnote{YOLOv5 question on overtaken YOLOv4 features: \url{https://github.com/ultralytics/yolov5/issues/370} (accessed 2021-04-28)} such as the CSPNet backbone \cite{wang2020cspnet}, and partly had their origin in prior YOLOv4 contributions by the YOLOv5 maintainer. One of the most notable data augmentation aspects is the mosaic loader in which four images are altered and combined to form a new image. This allows detection of objects outside of their normal context and at smaller sizes, which reduces the need for large mini-batch sizes. YOLOv5 reports high inference speed and small model sizes, allowing a convenient translation to mobile use cases via model export.

The approach on DFU detection via YOLOv5 described in the following is based on the early version v1.0\footnote{YOLOv5 v1.0: \url{https://github.com/ultralytics/yolov5/releases/tag/v1.0} (accessed 2020-09-12)} \cite{jocher2020a} commit \texttt{a1c8406}\footnote{YOLOv5 GitHub commit \texttt{a1c8406}: \url{https://github.com/ultralytics/yolov5/commit/a1c8406} (accessed 2020-08-29)} from 14 July 2020 that still exhibited several issues.

\subsubsection*{Pre-processing}

Initially, image data of the training dataset was analyzed via AntiDupl\footnote{AntiDupl GitHub repository: \url{https://github.com/ermig1979/AntiDupl} (accessed 2020-08-29)} in version \texttt{2.3.10} to identify duplicate images, yielding a set of 39 pair findings. A spatial analysis of duplicate pair annotation data was performed, utilizing the R language\footnote{R language website: \url{https://www.r-project.org/} (accessed 2020-08-29)} \cite{Rlang} in version \texttt{4.0.1} and the Simple Features for R (sf) package\footnote{Simple Features for R (sf) GitHub repository: \url{https://github.com/r-spatial/sf} (accessed 2020-08-29)} \cite{Rsf} in version \texttt{0.9-2}. Originally, none of the duplicate pair images showed bounding box intersections by themselves. After joining duplicate pair annotations, several intersections were detected with a maximum of two involved bounding boxes. These represented different annotations of the same wound in two duplicate images, now joint in one image. To resolve these, each intersection of two bounding boxes $\text{BBox}_1$ and $\text{BBox}_2$ were merged into a single bounding box $\widehat{\text{BBox}}$ by using their outer boundaries, as shown in Equ. \ref{eqn:bb_merging}.

\begin{equation}
	\label{eqn:bb_merging}
	\widehat{\text{BBox}} \;\;
	\begin{cases}
		\;\;\; \widehat{\text{xmin}} &= \;\; \min\big(\text{xmin}_1, \text{xmin}_2\big) \\
		\;\;\; \widehat{\text{ymin}} &= \;\; \min\big(\text{ymin}_1, \text{ymin}_2\big) \\
		\;\;\; \widehat{\text{xmax}} &= \;\; \max\big(\text{xmax}_1, \text{xmax}_2\big) \\
		\;\;\; \widehat{\text{ymax}} &= \;\; \max\big(\text{ymax}_1, \text{ymax}_2\big)
	\end{cases}
\end{equation}

The applied duplicate cleansing and annotation merging strategy resulted in $n = 1,961$ images with $k = 2,453$ annotations in the cleansed training dataset. Boundaries of merged bounding boxes were checked for consistency. Finally, annotation data was converted to the resolution-independent format used by YOLO implementations.

Reviewing image data of all dataset parts (training, validation and test), showed pronounced compression artifacts and color noise due to a high compression rate and downscaling to a low resolution. As both compression artifacts and color noise had derogatory effects on the detection performance, images were enhanced using a fast implementation of the non-local means algorithm \cite{buades2005} for color images, utilizing the Python language\footnote{Python language website: \url{https://www.python.org/} (accessed 2020-08-29)} in version \texttt{3.6.9} with the OpenCV on Wheels (opencv-python)\footnote{OpenCV on Wheels GitHub repository: \url{https://github.com/skvark/opencv-python} (accessed 2020-08-29)} package in version \texttt{4.2.0.34}. The algorithm parameters were set to \texttt{h} $= 1$ (luminance component filter strength) and \texttt{hColor} $= 1$ (color component filter strength) with \texttt{templateWindowSize} $= 7$ (template patch size in pixels) and \texttt{searchWindowSize} $= 21$ (search window size in pixels).

\begin{figure}[h!]\centering
	\centering
	\includegraphics[scale=0.6]{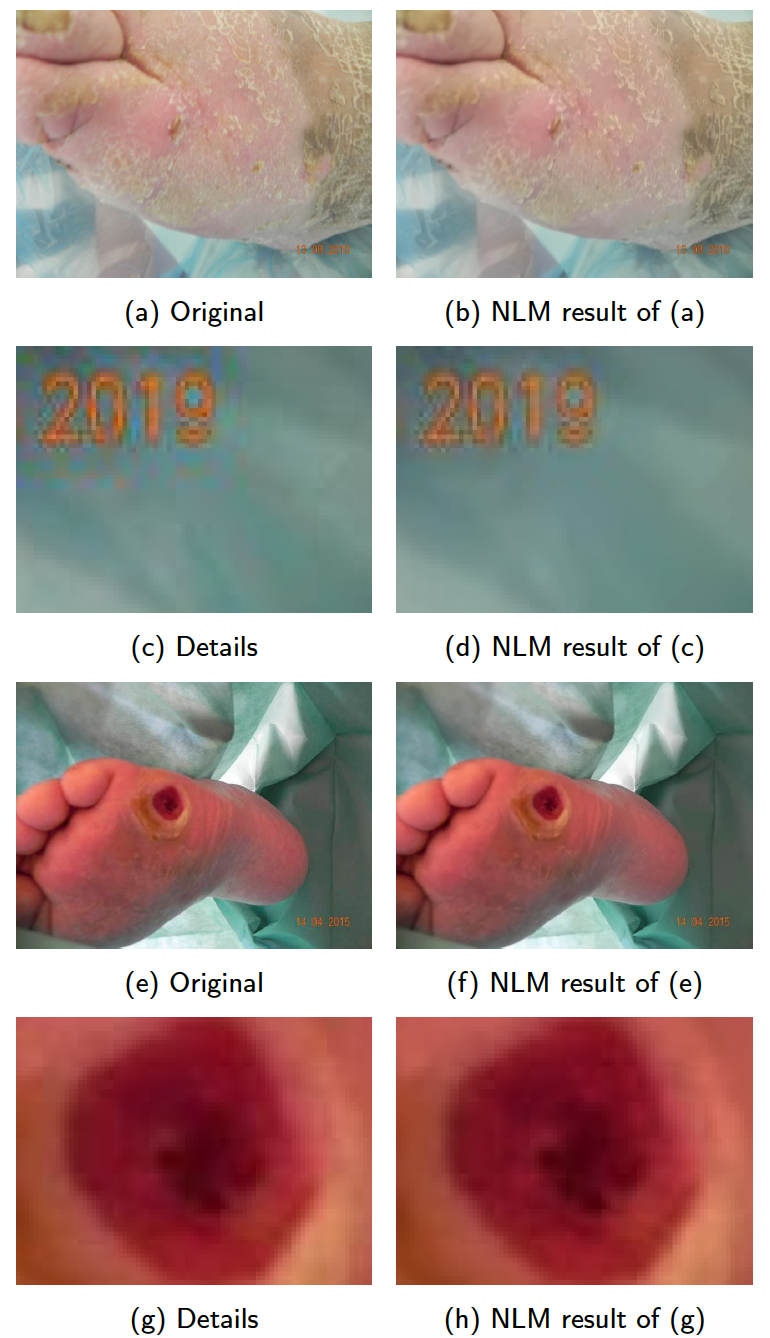}
	\caption{Effects of the non-local means (NLM) algorithm are shown for two example images (a) and (e) from the training dataset in (b) and (f). At a macroscopic level the changes are not obvious. At a detail level borders of compression artifacts on homogeneous areas and color noise of (c) are visibly reduced in (d). Vague textures of (g) are also more pronounced in (h).}
	\label{fig:original_nlm}
\end{figure}

Resulting images show less definitive compression artifact borders and notably reduced color noise. Some textures are also more pronounced. Examples of results at a macroscopic and a detail level are shown in Fig. \ref{fig:original_nlm}.

\subsubsection*{Data Augmentation}


YOLOv5 in v1.0 implements three sets of data augmentation techniques. The first set comprises alterations of colorspace components (hue, saturation, value), the second set comprises geometric distortions (random scaling, rotation, translation  and shearing), and the third set is represented by the mosaic loading of images.

A normalized fraction of $0.014$ images received hue augmentation, $0.68$ received saturation augmentation and $0.36$ received value augmentation. Scaling was applied in a normalized range of $\pm0.5$. Rotation, translation and shearing were disabled. Settings for colorspace component alterations and geometric distortions are definitions for distributions, generated during runtime by a random sampler for the augmentation function\footnote{YOLOv5 question on data augmentation: \url{https://github.com/ultralytics/yolov5/issues/2164} (accessed 2021-04-28)}. Using this approach, no image is presented more than once during training.

Mosaic data augmentation is comparable to CutMix, but takes four images instead of two and does not overlap them. Image parts are placed as quadrants in a new image with random ratios, thereby allowing the model to detect objects in different contexts and at different sizes. This reduces the need for large mini-batch sizes. However, the mosaic loader had to be disabled in the presented approach due to a bug, leading to invalid bounding boxes in resulting predictions.

\subsubsection*{Model}
\begin{figure}
	\centering
	\includegraphics[scale=0.58]{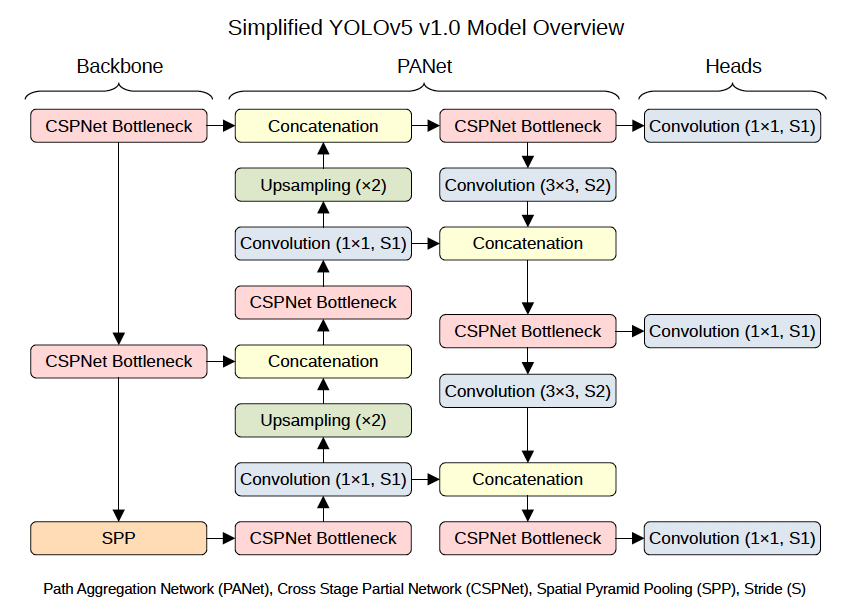}
	\caption{The architecture of YOLOv5 v1.0, adopted using \url{https://github.com/ultralytics/yolov5/issues/280} (accessed 2021-04-28), from community-driven discussions on the model representation, verified by the maintainer.}
	\label{fig:yolov5_v1_0_architecture}
\end{figure}

YOLOv5 includes four different models ranging from the smallest YOLOv5s with $7.5$ million parameters (plain $7$ MB, COCO pre-trained $14$ MB) and 140 layers to the largest YOLOv5x with $89$ million parameters and 284 layers (plain $85$ MB, COCO pre-trained $170$ MB). In the approach considered in this paper, the pre-trained YOLOv5x model is used. The general YOLOv5 v1.0 architecture is displayed in Fig. \ref{fig:yolov5_v1_0_architecture}. Different model sizes s, m, l and x vary in set depth and width factors for the model and its layer channels, which are $1.33$ and $1.25$ for the YOLOv5x model. 


The YOLOv5x model uses a detector that consists of a Cross Stage Partial Network (CSPNet) \cite{wang2020cspnet} backbone trained on MS COCO \cite{lin2014microsoft}, and a model head using a Path Aggregation Network (PANet) \cite{liu2018path} for instance segmentation. The backbone further incorporates a Spatial Pyramid Pooling (SPP) network \cite{he2015spatial}, which allows for dynamic input image size and is robust against object deformations. 

\subsubsection*{Training}
The hardware setup used for the experiment comprised a single NVIDIA\textregistered\ V100\footnote{NVIDIA\textregistered\ V100: \url{https://www.nvidia.com/en-us/data-center/v100/} (accessed 2020-08-30)} tensor core graphical processing unit (GPU) with 16 GB memory as part of an NVIDIA\textregistered\ DGX-1\footnote{NVIDIA\textregistered\ DGX-1: \url{https://www.nvidia.com/en-us/data-center/dgx-1/} (accessed 2020-08-30)} supercomputer for deep learning. YOLOv5 was set up using a provided Docker container\footnote{YOLOv5 Docker Hub container: \url{https://hub.docker.com/r/ultralytics/yolov5} (accessed 2020-08-30)}, executed via Nvidia-Docker\footnote{Nvidia-Docker GitHub repository: \url{https://github.com/NVIDIA/nvidia-docker} (accessed 2020-08-30)} in version \texttt{19.03.5}.

Training was organized in two stages: Initial training and self-training. The initial training stage uses the original available training data to train a model. The self-training approach, also called pseudo-labelling, extends available training data by inferring detections on images for which originally no annotation data is available \cite{koitka2017}. This is realized using the model resulting from the initial training stage; yielded detections are then used as pseudo-annotation data. Resuming the initial training in the self-training stage with the extended training data generalizes detection capabilities of the model.

A five-fold cross-validation was performed for each training stage to approximate training optima. Both stages used the default set of hyperparameters (including parameters related to the data augmentation procedures): \texttt{optimizer} $= \texttt{SGD}$, \texttt{lr0} $= 0.01$, \texttt{momentum} $= 0.937$, \texttt{weight\_decay} $= 0.0005$, \texttt{giou} $= 0.05$, \texttt{cls} $= 0.58$, \texttt{cls\_pw} $= 1.0$, \texttt{obj} $= 1.0$, \texttt{obj\_pw} $= 1.0$, \texttt{iou\_t} $= 0.2$, \texttt{anchor\_t} $= 4.0$, \texttt{fl\_gamma} $= 0.0$, \texttt{hsv\_h} $= 0.014$, \texttt{hsv\_s} $= 0.68$, \texttt{hsv\_v} $= 0.36$, \texttt{degrees} $= 0.0$, \texttt{translate} $= 0.0$, \texttt{scale} $= 0.5$, and \texttt{shear} $= 0.0$. A default seed value of $0$ was used for model initialization. Both training stages were performed in the single-class training mode, with mosaic data augmentation deactivated due to issues regarding bounding box positioning in the current YOLOv5 implementation.

During the initial training stage, a base model was trained on the pre-processed training dataset for $60$ epochs with a batch size of $30$. This base model was initialized with weights from the MS COCO pre-trained YOLOv5x model. For the self-training approach, the base model was then used to create the extended training dataset for self-training. Pseudo-annotation data was inferred for the validation and test datasets, using the best-performing epoch automatically saved at epoch 58. The resulting extended training dataset contained $4,161$ images, of which $3,963$ included $4,638$ wound annotations.

During the self-training stage, the base model training was resumed at its latest epoch, but trained further on the extended training dataset with a batch size of $20$. Three final training states were created: (1) after an additional $30$ epochs, (2) after an additional $40$ epochs, and (3) after an additional $60$ epochs of self-training (referred to as \texttt{E60\_SELF90}, \texttt{E60\_SELF100}, and \texttt{E60\_SELF120}).

\begin{figure*}
	\centering
	\includegraphics[scale=0.53]{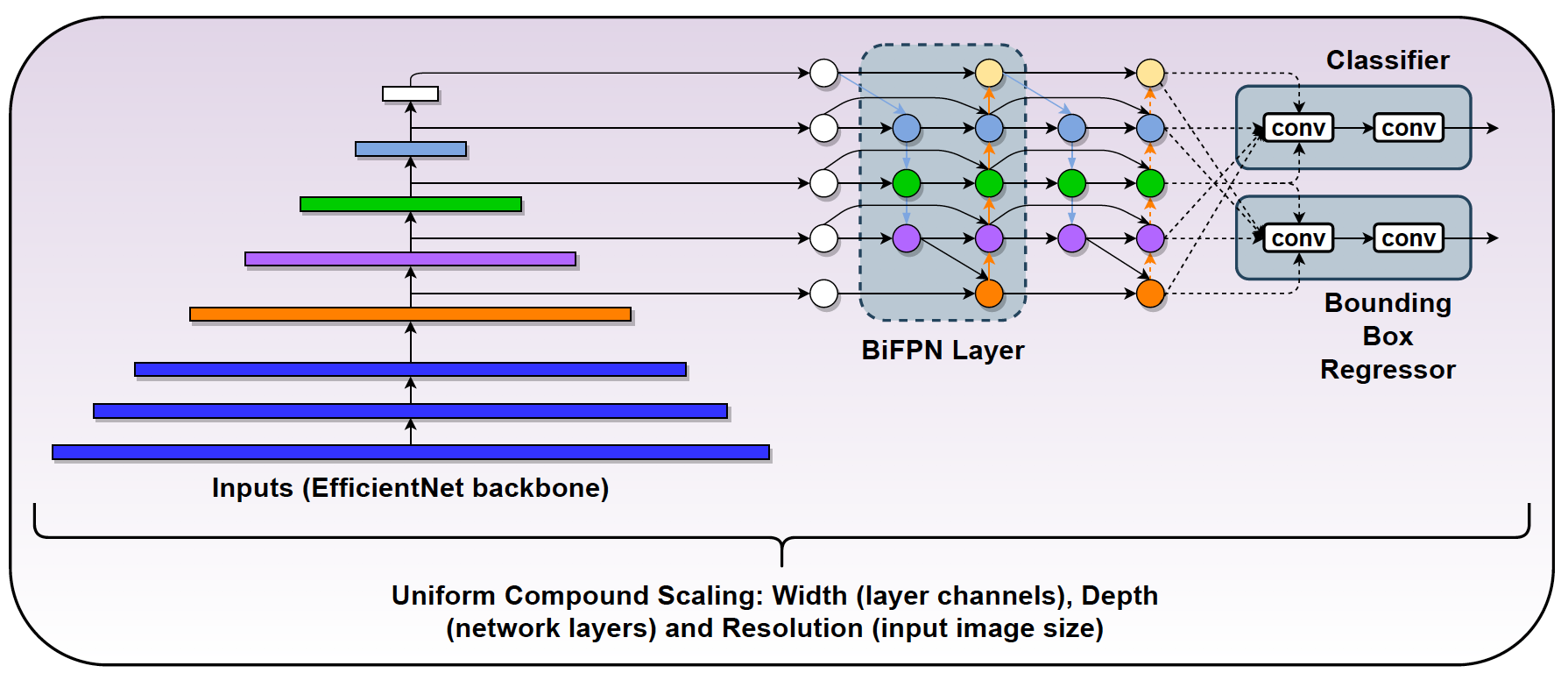}
	\caption{The architecture of EfficientDet \cite{tan2020efficientdet}.}
	\label{fig:effarch}
\end{figure*}

\subsubsection*{Post-processing}
The minimum confidence threshold for detection was set to $0.70$, so that only highly certain predictions were exported. This applies for pseudo-annotation data of the extended training dataset created for self-training as well as for the final predictions.

Predictions for our experiments were inferred via the final training states \texttt{E60\_SELF90}, \texttt{E60\_SELF100}, and \texttt{E60\_SELF120}, using the best epochs $88$, $96$ and $118$ respectively. An additional experiment was conducted based on the training state \texttt{E60\_SELF100} involving the built-in test-time augmentation and non-maxima suppression (NMS) features of YOLOv5 for inference.

Test-time augmentation (TTA) is a data augmentation method which involves several augmented instances of an image that are presented to the model. For each instance, predictions are made; these predictions for the image provide an ensemble of instance predictions. This can enable a model to detect objects it may not be able to detect in a ``clean" image. However, TTA may also cause multiple distinct detections for the same object that can harm evaluation scores. To tackle these, NMS was applied to collapse multiple intersecting detections into a single bounding box. The intersection over union (IoU) threshold was set low to $\text{IoU} \geq 0.30$, as in cases of multiple wounds in an image usually a distinct spatial demarcation was given. Thus, the risk of interfering detections of different wounds was low.

\subsection{EfficientDet} 
The EfficientDet architecture \cite{tan2020efficientdet} is an object detection network created by the Google Brain team, and utilises the EfficientNet ConvNet \cite{tan2019effnet} classification network as its backbone. EfficientDet uses feature fusion techniques in the form of a bidirectional feature pyramid network (BiFPN) which combines representations of input images at different resolutions. BiFPN adds weights to input features which enables the network to learn the importance of each feature. The outputs from the BiFPN are then used to predict the class of the detected object and to generate bounding boxes using bounding box regression. The main feature of EfficientDet is its ability to utilise compound scaling, which allows all parts of the network to scale in accordance to the target hardware being used for training and inference \cite{tan2020efficientdet}. An overview of the EfficientDet architecture is shown in Fig. \ref{fig:effarch}.

\subsubsection{Pre-processing}
\begin{figure}[htb!]
	\centering
	\includegraphics[scale=0.59]{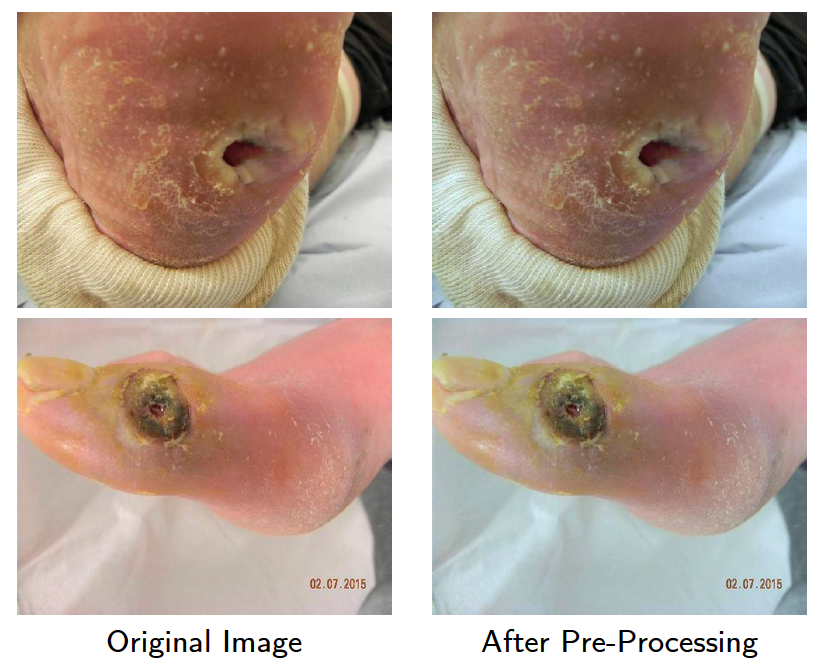}
	
	\caption{Shades of gray algorithm for pre-processing of the DFUC2020 dataset: (left) original images; and (right) pre-processed images.}
	\label{fig:prep}
\end{figure}

The dataset was captured with different types of camera devices under various lighting conditions. To counter variations in noise and lighting found in the dataset images, the Shades of Gray (SoG) color constancy algorithm was used \cite{hua2019effect}. Examples of pre-processed DFU images using SoG are shown in Fig. \ref{fig:prep}.

\subsubsection{Data Augmentation}
\begin{figure*}
	\centering
	\includegraphics[scale=0.59]{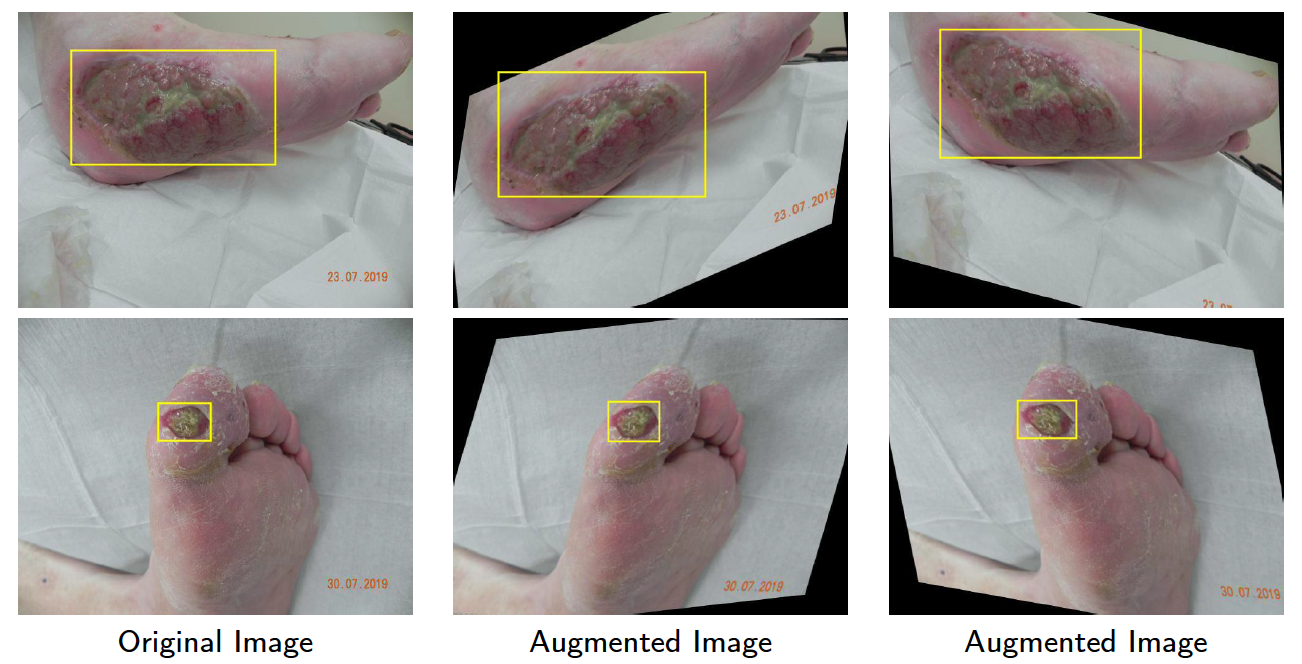}
	\caption{Bounding box data augmentation on the DFUC2020 dataset}
	\label{fig:dataaug}
\end{figure*}

Data Augmentation techniques have been proven to be an important tool in improving the performance of deep learning algorithms for various computer vision tasks \cite{goyal2018region, yap2020breast}. For the application of EfficientDet, we augmented the training data by applying identical transformations to the images and associated bounding boxes for DFU detection. Random rotation and shear transformations were used to augment the DFUC2020 dataset. Shearing involves the displacement of the image at its corners, resulting in a skewed or deformed output. Examples of these types of data augmentation are shown in Fig. \ref{fig:dataaug}.

\subsubsection{Model}
EfficientDet algorithms achieved state-of-the-art accuracy on the popular MS-COCO \cite{lin2014microsoft} object detection dataset. EfficientDet pre-trained weights are classed from D0 to D7, with D0 having the fewest number of parameters and D7 having the highest number of parameters. Tests on the MS-COCO dataset indicate that training using weights with more parameters results in better network accuracy. However, this comes at the cost of significantly increased training time. Given that the DFUC2020 dataset images were resized to 640x480, we selected the EfficientDet-D1 pre-trained weights for DFU detection \cite{goyal2020refined}.

\subsubsection{Training}
We trained the EfficientDet-D1 method on an NVIDIA Quadro RTX 8000 GPU (48GB) with a batch-size of 16, SGS optimizer with a learning rate of 0.00005, momentum of 0.9 and number of epochs set to 50. We used the validation accuracy with early stopping to select the final model for inference.

\subsubsection{Post-processing}
We further refined the EfficientDet architectures with a score threshold of 0.5 and removed overlapping bounding boxes to minimize the number of false positives. The scores were compared between the overlapping bounding boxes, with the bounding box with the highest score used as the final output.

\subsection{Cascade Attention DetNet} 

\subsubsection{Data Augmentation}
\begin{figure}[htp]
	\centering
	\includegraphics[scale=0.59]{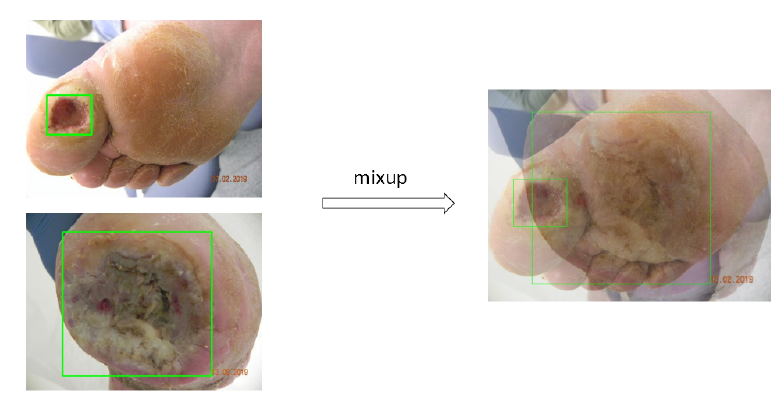}
	\caption{ The effect of the visually coherent image mixup method. }
	\label{fig:mixup}
\end{figure}

Given that the DFUC2020 dataset has only 2,000 images for training, we use two data augmentation methods to complement the dataset in order to avoid over-fitting when training models. A more generalized model can be obtained through data augmentation in order to make it adapt to the complex clinical environment. We use common data augmentation methods including horizontal and vertical image flipping, random noise and a central scaling method (which scales with ground truth as the center). Additionally, we increase the number of training images by using the visually coherent image mixup method \cite{mixup}. The original purpose of this method is to overcome the problem of disturbance rejection. Since Zhang et al. \cite{zhang2019bag} introduced this method into object detection, many researchers have used it in data augmentation to enhance network robustness. The principle of this algorithm involves the random selection of two sample images which are then used to generate a new sample image according to equation \ref{mixup1} and equation \ref{mixup2}.

\begin{equation}
	\label{mixup1}
	\hat{x} = \lambda x_i + (1-\lambda)x_j
\end{equation}
\begin{equation}
	\label{mixup2}
	\hat{y} = \lambda y_i + (1-\lambda)y_j
\end{equation}
where (\(x_i\), \(y_i\)), (\(x_j\), \(y_j\)) are the points of two sample images and $\lambda$ $\in$ [0,1], which is randomly generated by the Beta(alpha, alpha) distribution. The new sample (\(\hat{x}\), \(\hat{y}\)) is used for training. As shown in Fig. \ref{fig:mixup}, two images of DFU are mixed in a certain ratio. We use Beta(1.5,1.5) for the images' synthesis.

\begin{figure}[htp]
	\centering
	\includegraphics[scale=0.59]{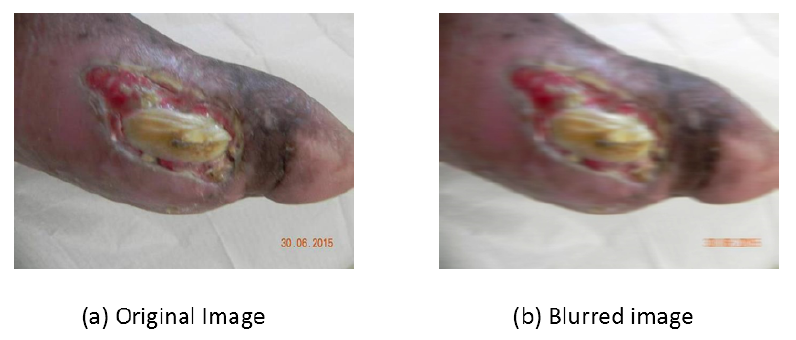}
	\caption{ The effect of the mobile fuzzy method. (a) shows the original image, and (b) shows the image after blurring with the mobile fuzzy method. }
	\label{fig:fuzzy}
\end{figure}

DFU detection can be challenging in complex environments, such as clinical settings, due to the large number of objects that might be present. To improve the ability of detection, we use the mobile fuzzy method for data augmentation, as shown in Fig. \ref{fig:fuzzy}.

\begin{figure*}
	\centering
	\includegraphics[scale=0.28]{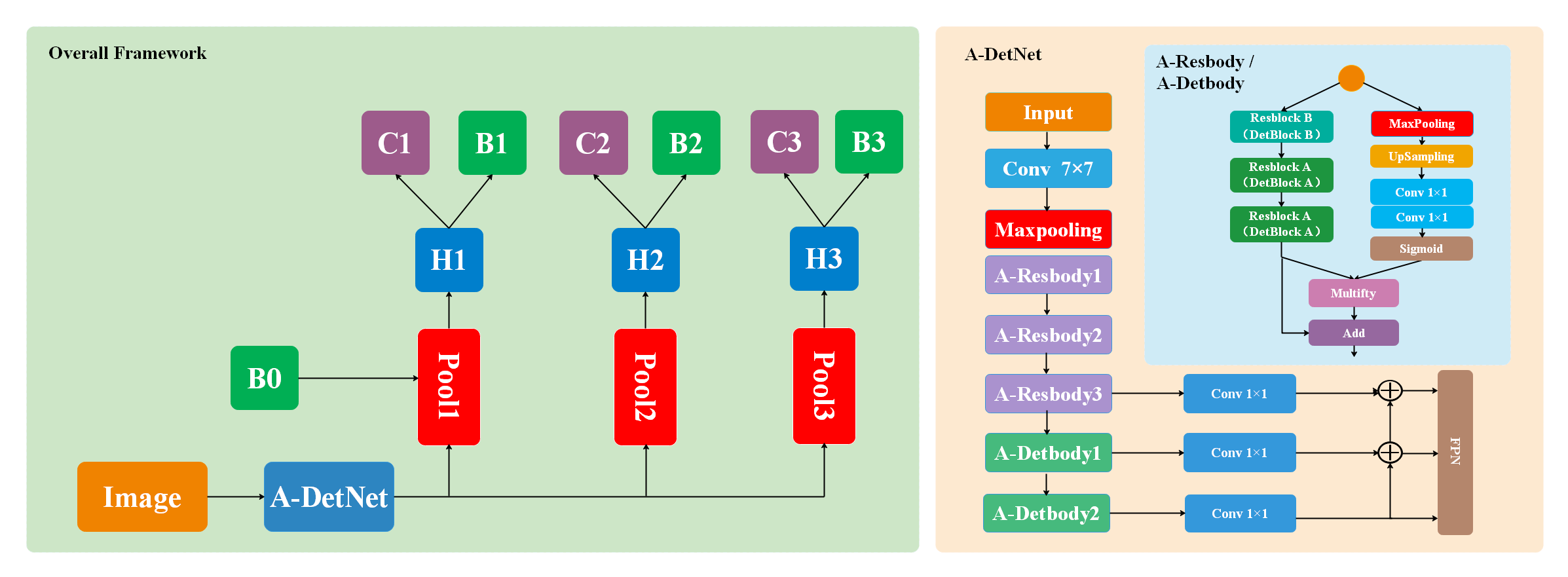}
	\caption{ The framework of CA-DetNet. ``Image" is an input image. ``A-DetNet" is a backbone network. ``Pool" represents region-wise feature extraction. ``H" is a network head. ``B" is a bounding box and ``C" represents classification. ``B0" is the proposal in all architectures. The structure of the A-DetNet is based on the DetNet. The attention mechanism is applied in Resbody and Detbody. Different bottleneck blocks in the Detbody or Resbody are similar to those in the DetNet. }
	\label{fig:cadetnet}
\end{figure*}

\subsubsection{Model}
The Cascade R-CNN \cite{Cai2017Cascade} is the first cascaded object detection model. Due to the superior performance of the cascade structure, it is widely used in the field of object detection \cite{zhao2020pointer}. We use the cascade structure in conjunction with DetNet \cite{2018DetNet}, which is designed to address the problems incurred by down-sampling repeatedly, as such a process reduces the accuracy of positioning. DetNet makes full use of the dilated convolutions to enhance the receptive field instead of down-sampling repeatedly. The overall framework of our method, Cascade Attention DetNet (CA-DetNet) is shown in Fig. \ref{fig:cadetnet}.

The detection of DFU is different from common object detection tasks. For common object detection tasks, objects can appear anywhere in the image. For the detection of DFU, the wounds can only appear on the foot, which is a good fit for applying an attention mechanism, which we added into the DetNet by adopting the mask branch of the Residual Attention Network \cite{2017Residual}.

The Attention DetNet (A-DetNet) is composed of 6 stages. The first stage consists of a 7 $\times$ 7 convolution layer (with a stride of 2) and a max-pooling layer. The second, third and fourth stages contain an A-Resbody, with the fifth and sixth stages containing an A-Detbody. The A-Resbody and A-Detbody are similar to those in the original DetNet. The difference between A-DetNet and the original DetNet is the addition of an attention branch into the Resbody and Detbody. The attention branch is similar to the mask branch of the Residual Attention Network, while we take other parts from the original Resbody or Detbody as the trunk. The attention branch of the Resbody comprise of two zoom structures, which consist of a max-pooling layer and an up-sampling layer, followed by two 1 $\times$ 1 convolution layers activated by sigmoid functions. 

Given that the five times down-sampling results in a feature map that is too small to recover the original size by up-sampled, we only add one zoom structure into the attention branch of the A-Detbody. The feature map from the trunk is multiplied by the mask from the attention branch. To avoid consuming the value of the feature and breaking the identity mapping, we refer to the Residual Attention Network and add one to the mask.

\subsubsection{Training}
For the cascade structure, we set the total number of cascade stages to 3, with the intersect over union (IOU) threshold set to 0.5, 0.6 and 0.7 for each of the three stages. During training we use DetNet pre-trained model, which has been trained on the ImageNet dataset, to accelerate model convergence. We train on one GPU (NVIDIA Tesla P100) for 60 epochs, with a batch size of 4 and a learning rate of 0.001. The learning rate decreases 10 times at the 10th epoch, and then decreases another 10 times at the 20th epoch. We optimize the model with the Adam optimizer.

\subsubsection{Post-processing} 
Noise from the external environment can lead to many low confidence bounding boxes. These bounding boxes will reduce the performance of the detector, so we adopt a special threshold suppression method to suppress bounding boxes with low thresholds except when the detector detects only one bounding box. We set the threshold to 0.5.

\section{Results and Analysis} 
We report and analyse the results obtained using the methods described above. The evaluation metrics are the number of true positives (TP), the number of false positives (FP), recall, precision, F1-Score and mAP, as described in the diabetic foot ulcer challenge 2020 \cite{cassidy2020dfuc2020}. For the common object detection task, mAP is used as the main evaluation metric. However, in this DFU task, miss-detection (a false negative) has potentially severe implications as it may affect the quality of life of patients. An incorrect detection (a false positive) could increase the financial burden on health services. Therefore, we regard F1-Score as equally important as mAP for performance evaluation. 

\subsection{Faster R-CNN} 
\label{sec51}
Table \ref{table:FRCNN} summarizes the quantitative results of pure Faster R-CNN, its variants, and the final ensemble model. From the table, the performance of pure Faster R-CNN is on par with Cascade R-CNN. In contrast, employing the Deformable convolution or PISA module significantly improves the performance. After we ensemble the model, we reduce FP substantially, with a reduction in TP also observed. Although the ensemble method improves the precision of DFU detection, it does not improve the overall score. Therefore, the best result is achieved by Deformable Faster R-CNN, with a mAP of 0.6940 and F1-Score of 0.7434.

\begin{table}
	\centering
	\small\addtolength{\tabcolsep}{2pt}
	\renewcommand{\arraystretch}{1.5}
	\caption{Faster R-CNN. The first row shows the results of pure Faster R-CNN, the second row shows the results of Cascade R-CNN, the third row shows the results of Faster R-CNN with Deformable Convolution v2, the fourth row shows the results of Faster R-CNN with Prime Sample Attention, and the last row shows the results of the ensemble method.}
	\label{table:FRCNN}
	\scalebox{0.80}{
		\begin{tabular}{ccccccc}
			\hline
			Method  &TP&FP& Recall    & Precision  & F1-Score   & mAP            \\ \hline\hline
			Faster &1512&683&0.7210  & 0.6888  & 0.7046    &0.6338\\
			Cascade &1483&649&0.7072  & 0.6956  & 0.7014   &0.6309\\
			Deform &\textbf{1612}&628&\textbf{0.7687}  & 0.7196  & \textbf{0.7434} &\textbf{0.6940}\\ 
			PISA &1495&444&0.7129  & 0.7710  & 0.7408    &0.6518\\ \hline
			Ensemble &1447&\textbf{394}&0.6900 & \textbf{0.7860} & 0.7349& 0.6353\\
			\hline
			\hline
	\end{tabular}}
\end{table}

\begin{figure}
	\centering
	\includegraphics[scale=0.51]{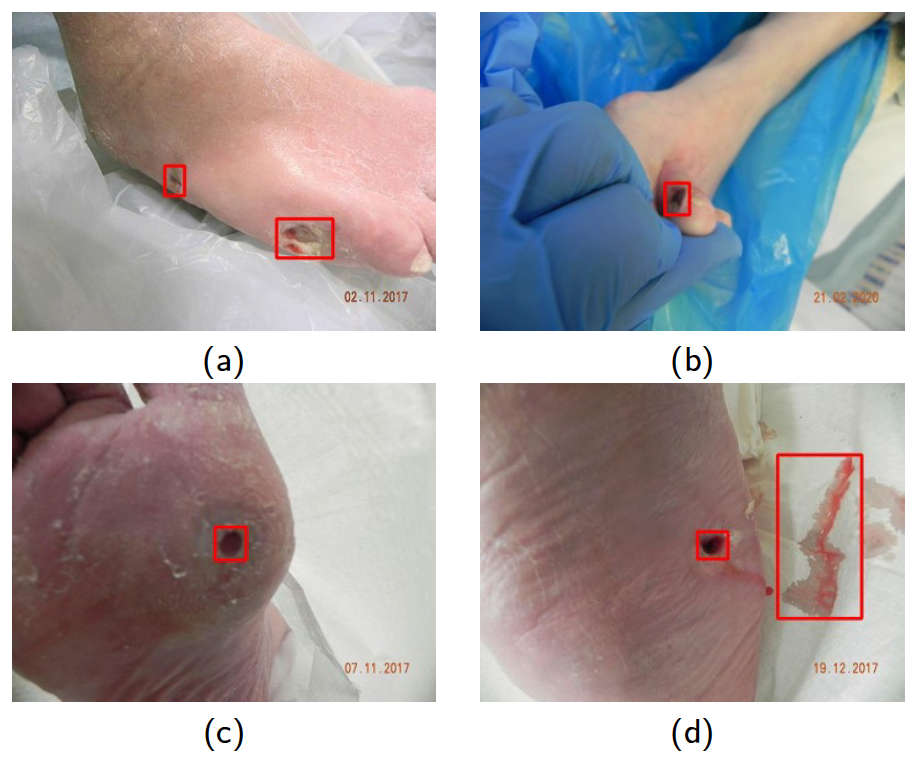}
	\caption{The qualitative results of Faster R-CNN with Deformable Convolution, which shows the best performance among Faster R-CNN based methods. It is noted that the network is able to detect small ulcers as shown in (a),(b) and (c). An example of a FP generated by the network is shown in (d).}
	\label{fig:faster_results}
\end{figure}

The qualitative results of Faster R-CNN with Deformable Convolution is summarized in Fig. \ref{fig:faster_results}. It can be seen that our model successfully detected the wounds in the image, even though the wounds are small (top-left, bottom-left and bottom-right images) or the images are blurred (top-right image). However, we observed the miss-detection as in the bottom-right image. In this image, the background texture of the blood was incorrectly detected as a DFU. To improve prediction accuracy, the training data should be captured in various environments so that the network is better able to discern between DFU and background objects.

\subsection{YOLOv3} 
Table \ref{table:YOLOv3} shows the final results of the proposed YOLOv3 method on the testing dataset. The results are reported for two different batch sizes, with and without post-processing.

As the results indicate, using a batch size of 50 leads to a better overall performance compared to using a batch size of 32. It also demonstrates that removing the overlaps leads to improvement in both F1-score and Precision, while resulting in slight decreases to both mAP and Recall. As the gain overpowers the loss, we conclude that removing overlaps results in better overall performance. 

\begin{figure}[!ht]
	\centering
	\includegraphics[scale=0.51]{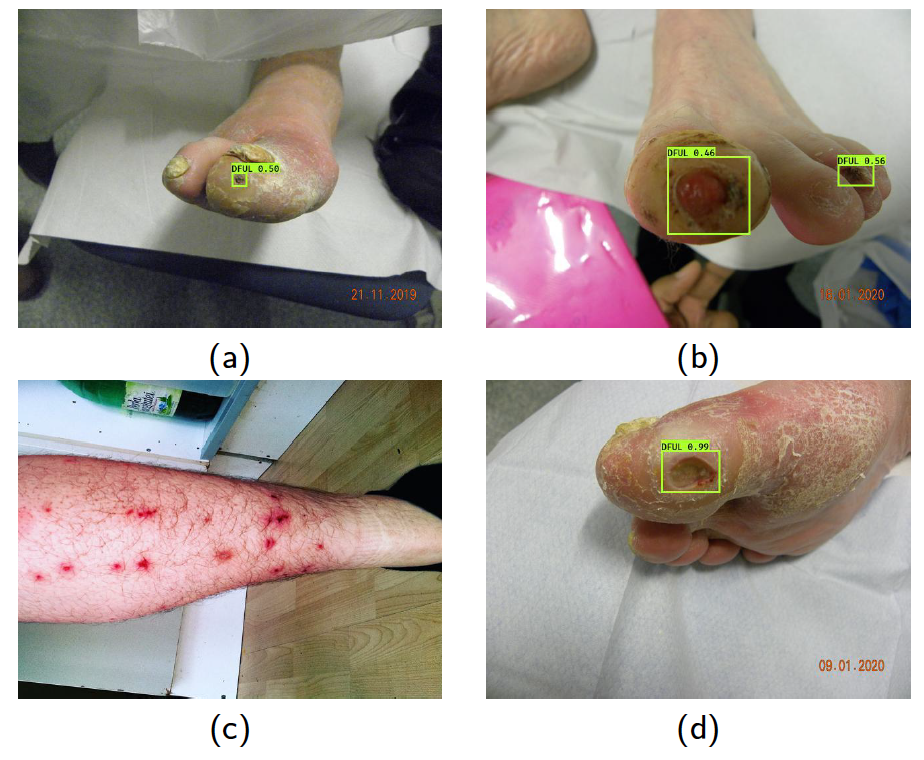}
	\caption{Examples of final detection outputs of trained YOLOv3, after post-processing.}
	\label{fig:yolo3_examples_final}
\end{figure}

While removing the detections with less than 0.3 confidence results in slightly better precision, it reduces recall, F1-score and mAP. Therefore, unless precision is the priority, removing the low confidence detections would not lead to an improvement. Examples of final detections for YOLOv3 are presented in Fig. \ref{fig:yolo3_examples_final}. 

Additionally, we added 60 copyright-free images of healthy feet\footnote{Website: \url{https://www.freepik.com/} (accessed 2020-08-29)} to the training set to observe the effect on detection performance. As shown in Table \ref{table:YOLOv3}, this results in an improvement of F1-Score, but reduces mAP.

\begin{table*}
	\centering
	\renewcommand{\arraystretch}{1.5}
	\caption{YOLOv3: Results of different settings, post-processing and adding extra copyright free foot images. B50 and B32: compares the performance of the method with batch size 50 with 32. Overlap-Removed: indicates the performance of the method, with overlap removal post processing. conf0.3: shows the impact of ignoring predictions with $<$ 0.3 confidence. Extra: demonstrates the effect on performance of adding extra images of healthy feet.}
	\scalebox{0.8}{
		\label{table:YOLOv3}
		\begin{tabular}{lccccccccc}
			\hline
			Method & \multicolumn{3}{l}{Settings}  & \multicolumn{6}{l}{Metrics} \\
			
			& Base & Coefficient & Overlap-Removed & TP & FP & Recall & Precision & F1-Score & mAP \\
			
			\hline\hline
			B50&50&0&$\times$ &\textbf{1572}&676& \textbf{0.7496}  & 0.6993  & 0.7236    & \textbf{0.6560}\\
			B50\_Overlap &50&0&\checkmark&1553&618&0.7406  & 0.7153  & 0.7277   &0.6500\\ 
			B32&32&0&$\times$ &1452&605&0.6929  & 0.7060  & 0.6994    &0.6053\\			
			B32\_Overlap &32&0&\checkmark&1433&551&0.6834  & 0.7223  & 0.7023   &0.5998\\
			B32\_Overlap\_conf &32&0.3&\checkmark&1386&\textbf{490}&0.6609  & \textbf{0.7388}  & 0.6977    &0.5835\\
			B50\_Exact &50&0&$\times$&1563&616&0.7454 & 0.7173  & 0.7311    &0.6548\\
			B50\_Overlap\_Extra &50&0&\checkmark&1543&565&0.7358 & 0.7320  & \textbf{0.7339}    &0.6484\\
			
			\hline
		\end{tabular}
	}
\end{table*}

\subsection{YOLOv5} 
\begin{figure}[h!]
	\centering
	\includegraphics[scale=0.59]{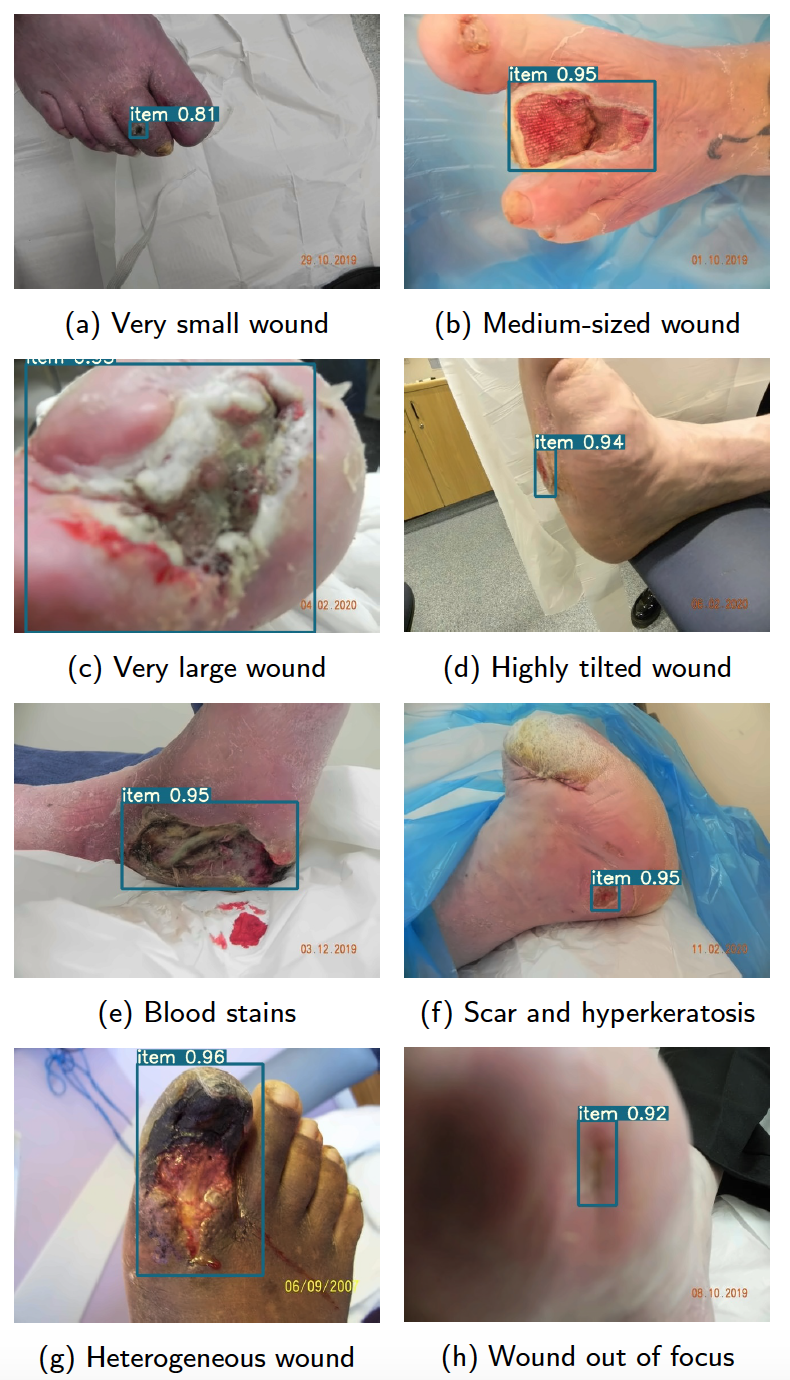}
	
	\caption{Examples for adequate predictions with YOLOv5 for different DFU sizes and compositions: (a) to (c) different wound sizes, (d) partially visible wound, (e) non-detected blood stain on dressing, (f) non-detected scar and hyperkeratosis, (g) heterogeneous wound composition, (h) detected wound out of focus.}
	\label{fig:yolov5_det_good}
\end{figure}

\begin{figure}[h!]	\centering
	\includegraphics[scale=0.59]{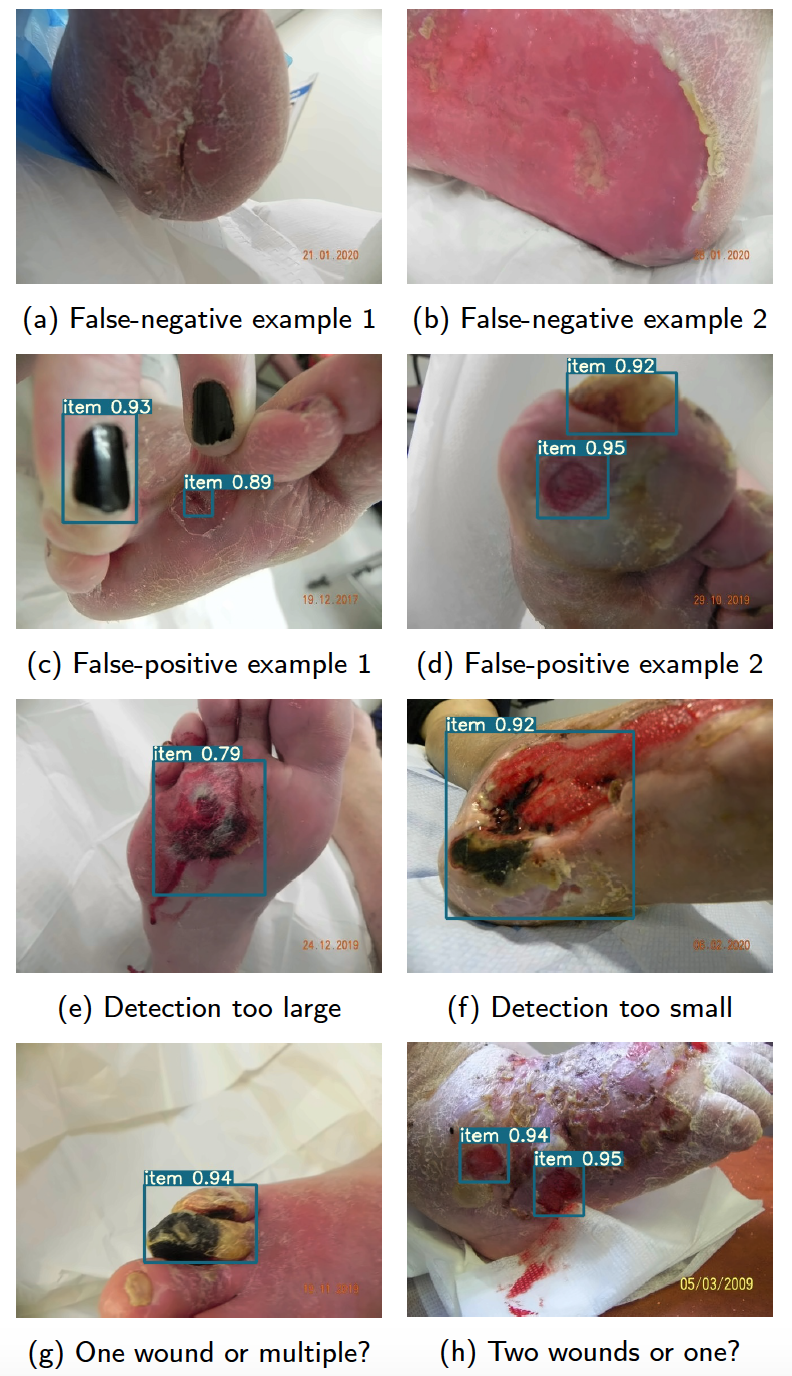}
	\caption{Examples of false negative, false positive, inadequate and questionable YOLOv5 predictions: (a) and (b) non-detected wounds, (c) and (d) painted finger nail and malformed toe nail, (e) and (f) too large and too small, (g) and (h) unclear detections (one, two, many?).}
	\label{fig:yolov5_det_bad}
\end{figure}

Table \ref{table:YOLOv5} summarizes the results of YOLOv5. Fewer additional self-training epochs in method E60\_SELF90 achieved better results than E60\_SELF100 and E60\_SELF120. However, the application of TTA with NMS on E60\_SELF100 achieved the best results in E60\_SELF100\_TTA\_NMS. Examples of detections with E60\_SELF100\_TTA\_NMS on the test set are shown in Fig. \ref{fig:yolov5_det_good}, Fig. \ref{fig:yolov5_det_bad} shows additional examples of false negative and false positive cases.

\begin{table*}
	\centering
	\renewcommand{\arraystretch}{1.5}
	\caption{YOLOv5: Results of different submitted runs. The settings state epochs for base and self-training as well as the use of test-time augmentation (TTA) and non-maximum suppression (NMS). Best results are highlighted bold, the winning method is highlighted gray.}
	\scalebox{0.8}{
		\label{table:YOLOv5}
		\begin{tabular}{lccccccccc}
			\hline
			Method & \multicolumn{3}{l}{Settings}  & \multicolumn{6}{l}{Metrics} \\
			
			& Base & Self-training & TTA+NMS & TP & FP & Recall & Precision & F1-Score & mAP \\
			\hline
			\hline
			E60\_SELF90 & 60 & 30 & $\times$ & 1504 & \textbf{474} & 0.7172 & \textbf{0.7604} & \textbf{0.7382} & 0.6270 \\
			E60\_SELF100 & 60 & 40 & $\times$ & 1496 & 485 & 0.7134 & 0.7552 & 0.7337 & 0.6165 \\
			E60\_SELF100\_TTA\_NMS & 60 & 40 & \checkmark & \textbf{1507} & 498 & \textbf{0.7187} & 0.7516 & 0.7348 & \textbf{0.6294} \\
			E60\_SELF120 & 60 & 60 & $\times$ & 1502 & 478 & 0.7163 & 0.7586 & 0.7368 & 0.6201 \\
			\hline
		\end{tabular}
	}
\end{table*}

\subsection{EfficientDet} 
\begin{figure}[!t]
	\centering
	\includegraphics[scale=0.59]{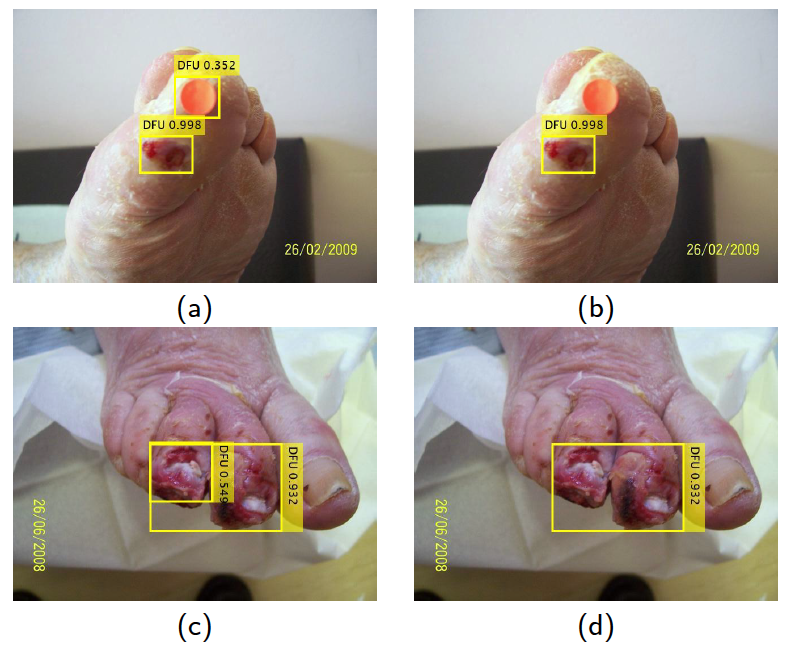}
	\caption{The results of EfficientDet. (a) and (c) are the results of EfficientDet without post-processing; (b) and (d) are the results obtained with post-processing.}
	\label{fig:fp}
\end{figure}

Table \ref{table:EffDet} shows the results of the EfficientDet model on the DFUC2020 testing set both with and without post-processing. The results indicate that the number of both TP and FP cases are reduced with the post-processing method. However, with the post-processing method, the percentage of TP cases (from 1,626 to 1,593) is 2.02\% compared to FP cases (from 720 to 594), which is 17.50\%. Hence, the post-processing method results in an important improvement in both Precision (67.86\% to 72.84\%) and F1-score (72.38\% to 74.37\%), with a slight decrease in both mAP (57.82\% to 56.94\%) and Recall (77.44\% to 75.97\%). The EfficientDet with post-processing method achieved the highest F1-Score and Precision (least number of FP cases) in DFUC2020. Examples of final outputs by the refined EfficientDet architecture are shown in Fig. \ref{fig:fp}. 

\begin{table}
	\centering
	\small\addtolength{\tabcolsep}{2pt}
	\renewcommand{\arraystretch}{1.5}
	\caption{EfficientDet. `Before' is the result of EfficientDet without post-processing and `After' is the result with post-processing.}
	\label{table:EffDet}
	\scalebox{0.75}{
		\begin{tabular}{ccccccc}
			\hline
			Methods  &TP&FP& Recall    & Precision  & F1-Score   & mAP            \\ \hline\hline
			Before &\textbf{1626}&770&\textbf{0.7754} & 0.6786  & 0.7238    &\textbf{0.5782}\\ 
			After &1593&\textbf{594}&0.7597 & \textbf{0.7284}  & \textbf{0.7437}    &0.5694\\ 
			\hline
			\hline
	\end{tabular}}
\end{table}

\subsection{Cascade Attention DetNet}  
Table \ref{table:Cascade Attention DetNet} summarizes the results of the Cascade Attention DetNet on the DFUC2020 testing dataset. The results are reported for two different data augmentation methods, two different backbones and with or without a pre-trained model.

From the results, we observe that CA-DetNet with two data augmentation methods and the pre-trained model achieves the best result. It achieves the highest score of 63.94\% on mAP and 70.01\% on F1-Score. The C-DetNet achieves the highest score of 74.11\% on Recall, while the CA-DetNet with the mobile fuzzy method achieves the highest score of 66.67\% on Precision.

\begin{table*}
	\centering
	\renewcommand{\arraystretch}{1.5}
	\caption{Results for each of the Cascade Attention DetNets.}
	\scalebox{0.8}{
		\begin{tabular}{l|lll|llllll}
			Backbone & \multicolumn{3}{l}{Settings}  & \multicolumn{6}{l}{Metrics} \\ 
			\cline{2-10}
			&    pre-trained & mobile fuzzy & mixup &TP&FP &Recall&Precision& F1-Score & mAP \\
			C-DetNet &  $\checkmark$ & $\checkmark$  & $\checkmark$   &1554&789&\textbf{0.7411} & 0.6633  & 0.7000    &0.6391  \\
			CA-DetNet & $\times$ & $\times$  & $\times$   &1493&1089&0.7120  & 0.5782  & 0.6382    &0.5963  \\
			CA-DetNet & $\checkmark$ & $\times$  & $\times$   &1523&820&0.7263  & 0.6500  & 0.6860    &0.6204  \\
			CA-DetNet &$\checkmark$& $\times$   & $\checkmark$  &1431&961&0.6824  & 0.5982  & 0.6376    &0.5749  \\
			CA-DetNet & $\checkmark$ & $\checkmark$ & $\times$   &1528&764&0.7287  & \textbf{0.6667} & 0.6963    &0.6350  \\
			CA-DetNet &$\checkmark$ & $\checkmark$ & $\checkmark$ &1554&788&0.7411 & 0.6635  & \textbf{0.7002} &\textbf{0.6394}  \\ \hline
		\end{tabular}
	}
	\label{table:Cascade Attention DetNet}
\end{table*}

From the analysis, we observe that the mobile fuzzy data augmentation method brings about a striking effect and improves 1.46\% on mAP and 1.03\% on F1-Score. However, we note that using the single mixup method in data augmentation did not enhance the performance. The results suggest that the mobile fuzzy method allows the model to adapt to the noise from the external environment, while the mixup method is detrimental. The attention mechanism contributes to the improved performance of detection and increases mAP by 0.02\% and F1-Score by 0.03\%. Moreover, training with a pre-trained model can accelerate the convergence of the model and improve its ability to detect DFU.

\begin{figure}[htp]
	\centering
	\includegraphics[scale=0.59]{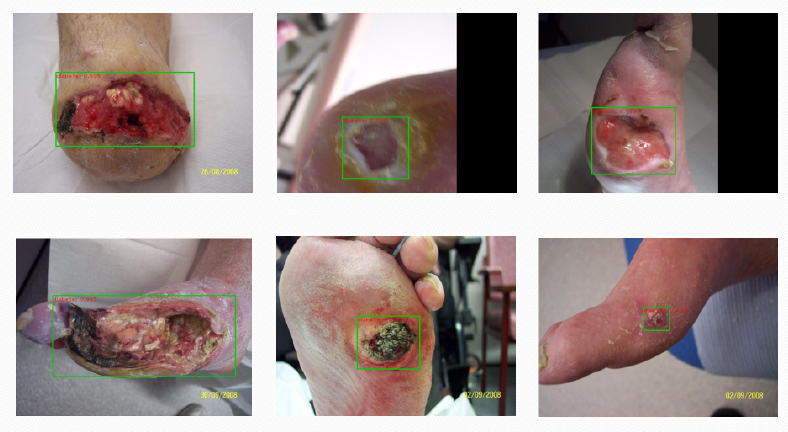}
	\caption{The results of CA-DetNet: Illustration of successful DFU detections. Note the variety of DFU sizes detected by the network, ranging from small (bottom-right), medium (top-middle and bottom-middle), large (top-left, top-right) and very large (bottom-left).}
	\label{fig:displayDFU}
\end{figure}

Our approach was effective for the vast majority of the detected cases, as shown in Fig. \ref{fig:displayDFU}. However, due to the visual complexity of clinical environments, there are also some failure cases in our approach. From our observations, such failures are generally due to the false identification of toenails, interference from the external environment and low image quality. For the false identification of toenails, we believe that the appearance of leuconychia is similar to wounds and some cases of DFU are located on or around the toenail. Background objects may also sometimes interfere with detection results. We use the attention mechanism to deal with this problem to some extent. For image quality, we observe that there are several images which are blurry. We use data augmentation methods like the mobile fuzzy method to partially address this problem. We speculate that a two-stage architecture with an initial stage to detect and segment the relevant foot area could be used to address this issue. However, additional labeled data may be required be achieve this goal.

\subsection{Comparison} 
The results from the popular deep learning object detection methods and the proposed CA-DetNet are comparable. Table \ref{table:bestmAP} shows the overall results when evaluated on the DFUC2020 testing set, where we present the best mAP from each object detection method. Considering the ranking based on mAP, the best result is achieved by the variant of Faster R-CNN using Deformable Convolution, with 0.6940. This method achieves the highest TP and the best Recall. It is noted that YOLOv5 achieved the lowest number of FP, but it has lower mAP and F1-Score. 

\begin{table}
	\centering
	\small\addtolength{\tabcolsep}{2pt}
	\renewcommand{\arraystretch}{1.5}
	\caption{A summary based on the mAP ranking from each object detection method when evaluated on the DFUC2020 testing set.}
	\label{table:bestmAP}
	\scalebox{0.75}{
		\begin{tabular}{ccccccc}
			\hline
			Methods  &TP&FP& Recall    & Precision  & F1-Score   & mAP            \\ \hline\hline
			Faster R-CNN &\textbf{1612}&628&\textbf{0.7687} &0.7196 & 0.7434& \textbf{0.6940}\\	
			YOLOv3 &1572&676& 0.7496 & 0.6993  & 0.7236    & 0.6560\\
			CA-DetNet&1554&788&0.7411 & 0.6635  & 0.7002 &0.6394  \\
			YOLOv5 & 1507 & \textbf{498} & 0.7187 & \textbf{0.7516} & 0.7348 & 0.6294\\ 
			EfficientDet &1593&594&0.7597 & 0.7284  & \textbf{0.7437}    &0.5694\\ 
			\hline
			\hline
	\end{tabular}}
\end{table}

In Table \ref{table:bestF1}, the ranking according to F1-Score shows the highest F1-Score of 0.7437 obtained by EfficientDet, however, this network reports the lowest mAP at 0.5694. On the other hand, the Faster R-CNN approach achieves a comparable F1-Score of 0.7434 with the highest mAP of 0.6940.

\begin{table}
	\centering
	\small\addtolength{\tabcolsep}{2pt}
	\renewcommand{\arraystretch}{1.5}
	\caption{A summary based on F1-Score ranking from each object detection method when evaluated on the DFUC2020 testing set.}
	\label{table:bestF1}
	\scalebox{0.75}{
		\begin{tabular}{ccccccc}
			\hline
			Methods  &TP&FP& Recall    & Precision  & F1-Score   & mAP            \\ \hline\hline
			EfficientDet &1593&594&0.7597 & 0.7284  & \textbf{0.7437}    &0.5694\\ 
			Faster R-CNN &\textbf{1612}&628&\textbf{0.7687} &0.7196 & 0.7434& \textbf{0.6940}\\	
			YOLOv5 & 1504 & 474 & 0.7172 & \textbf{0.7604} & 0.7382& 0.6270 \\
			YOLOv3 &1543&\textbf{565}&0.7358 & 0.7320  & 0.7339  &0.6484\\
			CA-DetNet&1554&788&0.7411 & 0.6635  & 0.7002 &0.6394  \\
			\hline
			\hline
	\end{tabular}}
\end{table}

\begin{figure}
	\centering
	\includegraphics[scale=0.59]{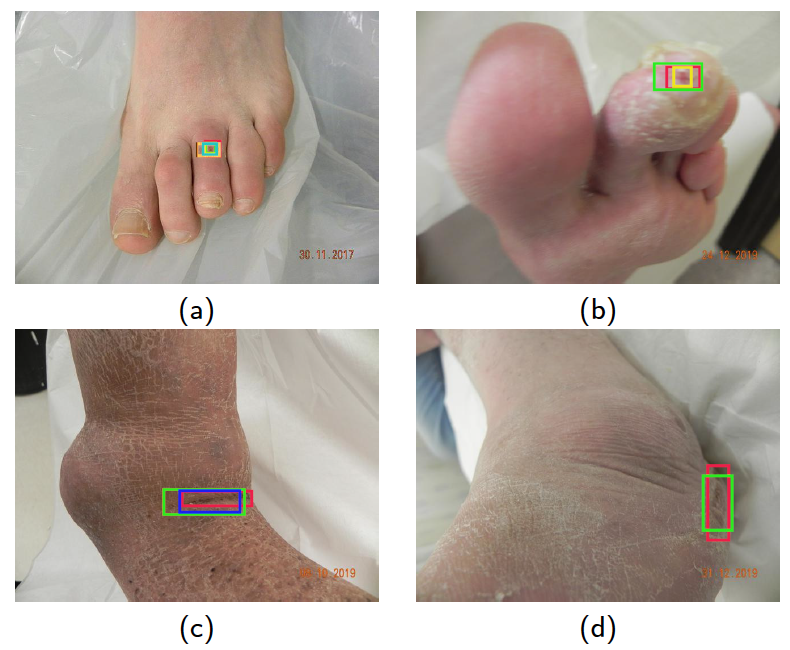}
	\caption{Visual comparisons of object detection methods when compared to the ground truth (in red): (a) An easy case where all the methods detected the ulcer; (b) A more challenging case detected by Faster R-CNN (green) and EfficientDet (yellow); (c) A challenging case detected by Faster R-CNN (green) and CA-DetNet (blue); and (d) A challenging case only detected by Faster R-CNN (green).}
	\label{fig:visual_comparison}
\end{figure}

Fig. \ref{fig:visual_comparison} visually compares the detection results on DFUs with less visible appearances. In Fig. \ref{fig:visual_comparison}(a), the ulcer was detected by all the methods. However, in Fig. \ref{fig:visual_comparison}(b), only Faster R-CNN and EfficientDet detected the ulcer. Fig. \ref{fig:visual_comparison}(c) is another challenging case and was detected by CA-DetNet and Faster R-CNN. In Fig. \ref{fig:visual_comparison}(d), we demonstrate a case where only Faster R-CNN successfully localised the ulcer.

In Section \ref{sec51}, we demonstrate that the ensemble method using Weighted Boxes Fusion did not improve the results of four Faster R-CNN approaches. This observation suggests that additional experiments based on different deep learning approaches should be investigated. We ran experiments based on combinations of two approaches (Faster R-CNN + (CA-DetNet / EfficientDet / YOLOv3 / YOLOv5)), three approaches and a combination of all approaches, as summarised in Table \ref{table:ensemble}. From our observation, the ensemble methods reduces the number of TPs and FPs, i.e., the more approaches used, the lower the number of TPs and FPs. This approach did not improve mAP, but in the majority of the ensembles there are notable improvements in precision, hence led to an improvement in F1-Score. The best F1-Score for the ensemble method is 0.7617, achieved by ensembling Faster R-CNN with Deformable Convolution and EfficientDet. 


Apart from fine-tuning each deep learning method to achieve maximum performance, the methods are highly dependent on the pre-processing stage, selection of data augmentation, post-processing methods and ensemble method. We address the limitations and future challenges of our work in the following section.

\begin{table}
	\centering
	\small\addtolength{\tabcolsep}{2pt}
	\renewcommand{\arraystretch}{1.5}
	\caption{A Comparison of ensemble methods with different combinations of object detection frameworks, where FRCNN is Faster R-CNN, DetNet is CA-DetNet, EffDet is EfficientDet and `ALL methods' represents an ensemble method based on Faster R-CNN, CA-DetNet, EfficientDet, YOLOv3 and YOLOv5.}
	\label{table:ensemble}
	\scalebox{0.66}{
		\begin{tabular}{ccccccc}
			\hline
			Methods  &TP&FP& Recall    & Precision  & F1-Score   & mAP            \\ \hline\hline
			FRCNN+DetNet &\textbf{1510} & 426 & \textbf{0.7201} & 0.7800 & 0.7488& \textbf{0.6619}\\
			FRCNN+EffDet & 1502 & 345 &0.7163 & 0.8132  & \textbf{0.7617}  &0.6425\\
			FRCNN+YOLOv3 &1423&310&0.6786 &0.8211 & 0.7431& 0.6205\\
			FRCNN+YOLOv5 &1453&350&0.6929 &0.8059 & 0.7451& 0.6421\\	
			FRCNN+YOLOv5+EffDet &1396&252&0.6657 &0.8471 & 0.7455& 0.6109\\
			FRCNN+YOLOv5+DetNet &1384&295&0.6600&0.8243 & 0.7331& 0.6132\\
			FRCNN+DetNet+EffDet & 1435 & 270 & 0.6843 & 0.8416 & 0.7549 & 0.6229 \\
			ALL methods &1277&\textbf{198}&0.6090 & \textbf{0.8658}  & 0.7150    &0.5642\\ 
			\hline
			\hline
	\end{tabular}}
\end{table}

\section{Discussion} 
In this section, we discuss the performance of each object detection method and future work to improve DFU detection. Whilst most of the results show an F1-Score $>$70\%, there are many challenges ahead to enable the use of deep learning algorithms in real-world settings.

Faster R-CNN based approaches detected DFU in the DFUC2020 testing set with high mAP and F1-Score. In addition, the variants of Faster R-CNN largely improve the performance of the original Faster R-CNN. After ensembling the results of four models, we reduced the number of false positives, however, the overall performance was less than the individual variants of Faster R-CNN. The reason may be that even though we are fusing the prediction of four models into one prediction, similar results are predicted among these four models because all models are based on Faster R-CNN. Therefore, in future work, a one-stage object detection method such as CenterNet \cite{zhou2019objects} could potentially be included in the ensemble method to produce more accurate results. 


The YOLOv3 algorithm is able to reliably detect DFU and ranked third place in both mAP and F1-Score ranking. We have observed that post-processing (by removing overlaps), along with the removal of low confidence detections, leads to an improvement in precision but at the expense of the number of true positives and recall. Additionally, our analysis indicates that adding additional images of healthy feet, along with post-processing, can result in a higher F1-score. We aim to further investigate the results of pre-processing, as well as studying a more effective post-processing method. 


The YOLOv5 approach demonstrated reliable detection performance with an overall high precision over the different model configurations. Application of the NLM algorithm for image enhancement and generalization via self-training helped to further increase precision. Improvements via duplicate cleansing and bounding box merging were marginal due to the limited number of cases, but could prove beneficial on larger datasets. Use of TTA with NMS further increased true-positive detections at the cost of increased false-positive cases, yet increased the mAP and F1-Score. For the presented approach, several optimizations may be possible. The least self-trained model performed best, indicating that models with less self-training epochs may perform better. Model Ensembling\footnote{YOLOv5 GitHub repository tutorial on Model Ensembling: \url{https://github.com/ultralytics/yolov5/issues/318} (accessed 2020-09-28)} could allow further performance improvements when fusing different specialized models. In addition, investigation of Hyperparameter Evolution\footnote{YOLOv5 GitHub repository tutorial on Hyperparameter Evolution: \url{https://github.com/ultralytics/yolov5/issues/607} (accessed 2020-09-28)} allows general hyperparameter optimization, given the required resources.

As the presented results were obtained with the initial release v1.0 \cite{jocher2020a} of YOLOv5, the resulting performance is limited compared to that achievable with matured up-to-date versions of the network \cite{bruengel2021detryolov5}. Since its release, YOLOv5 has been improving rapidly and its full potential could not be taken advantage of during the DFUC2020. E.g., in v1.0 the novel Mosaic data augmentation method was not functioning correctly on custom data. At the time of writing, the matured version v5.0\footnote{YOLOv5 v5.0: \url{https://github.com/ultralytics/yolov5/tree/v5.0} (accessed 2021-04-26)} \cite{jocher2021} is available, featuring numerous bug fixes, improvements and novelties. E.g., in the meantime the activation function changed from Leaky ReLU \cite{maas2013leaky_relu} in v1.0 (used here) to Sigmoid Linear Unit (SiLU) \cite{hendrycks2020gaussian} (since v4.0) \cite{glenn_jocher_2021_4418161}, further increasing detection performance. Due to its reasonable performance and mobile-focus, YOLOv5 will prove helpful when performing DFU detection tasks directly on mobile devices.


The refined EfficientDet algorithm is able to detect DFU with a high recall rate. The pre-processing stage using the Shades of Gray algorithm improved the color consistency of the images in the dataset. We extensively used data augmentation techniques to learn the subtle features of DFUs of various sizes and severity. The post-processing stage we implemented has refined the inference of the original EfficientDet method by removing overlapping bounding boxes. Due to low mAP, further work will focus on investigating the larger EfficientDet network architectures, particularly EfficientDet-D7.


The performance of Cascade Attention DetNet on the DFUC2020 testing set is not entirely satisfactory. We evaluated our model on 10\% of the DFUC2020 training set and it achieved an mAP of 0.9. We analyzed the possible reasons and speculate that the model may be over-fitting, to which ensemble learning may provide a possible solution. We further aim to use appropriate data augmentation methods to improve the robustness of the model.

The ensemble methods based on the fusion of different backbones have reduced the number of predicted bounding boxes substantially. Faster R-CNN with Deformable Convolution predicted 2,240 bounding boxes. However, after ensembling with EfficientDet, only 1,847 bounding boxes were predicted. The number of predicted bounding boxes dropped to 1,475 when we ensembled the results from all five networks. Consequently, the ensemble methods reduced the number of TPs and FPs. It is crucial for future research to focus on true positives, i.e. correctly locate the DFUs. One of the aspects required to overcome this issue is to understand the threshold setting of IoU. Our experiments used IoU $\geq$ 0.5, which is the guideline set by object detection for natural objects. However, in medical imaging studies \cite{drukker2002computerized, yap2008novel}, they used an IoU (or Jaccard Similarity Index) threshold of 0.4. When we evaluated the performance of the best ensemble method, the number of TPs increased to 1,594, with IoU $\geq$ 0.3 the number of TPs increased to 1,668. With Faster R-CNN with Deformable Convolution, the number of TPs increased to 1,743 and 1,883 for IoU thresholds of 0.4 and 0.3, respectively.

Currently, clinicians (podiatrists and diabetes consultants) visually assess the diabetic foot for detection of ulcers, taking photographs at the diagnosis stage and periodically re-evaluating on subsequent patient clinic appointments. This provides accurate assessment of wound healing progress by visually comparing photographs of ulcers at different stages of the disease. We have developed AI algorithms so that in the near future, patients can use mobile devices in their homes so that detection, assessment of wound progress and prognostication can be completed remotely without the need for frequent appointments to foot clinics. Our research in evaluating the performance of different deep learning frameworks in DFU detection is a crucial step to support future development in this field.

\section{Conclusion} 
We conduct a comprehensive evaluation of the performance of deep learning object detection networks for DFU detection. While the overall results show the potential of automatically localising the ulcers, there are many false positives, and the networks struggle to discriminate ulcers from other skin conditions. A possible solution to address this issue might be to introduce a second classifier in the form of a negative dataset to train future networks on. However, in reality, it may prove impossible to gather all possible negative examples for supervised learning algorithms. This approach could also impact network size and complexity, which could negatively impact inference speed. Segmenting the foot from its surroundings might provide another possible solution to this problem, so that trained models do not have to account for objects in complex environments. Other future research challenges include:
\begin{itemize}
	\item Increasing the size of the existing dataset with clinical annotations which would include metadata indicating the development stage of each DFU. However, in the real-world, there are still barriers in data sharing and clinical annotation is expensive and time consuming. It will be important to encourage the co-creation of such datasets via machine learning and clinical experts to foster a better understanding of the annotated data. While increasing the number of images may benefit the training process, other aspects such as ulcer location and image capture from subjects with various skin tones should be considered.
	\item Create self-supervised and unsupervised deep learning algorithms for DFU detection. These methods were developed and implemented for natural object detection tasks and remain under-explored in medical imaging.
	\item For inspections of DFU, accurate delineation of an ulcer and its surrounding skin can help to measure the progress of the ulcer. Goyal et al. \cite{goyal2017fully} developed an automated segmentation algorithm for DFU. However, they experimented on a small dataset and future work will potentially enable a larger scale of experimentation. 
	\item The use of DFU classification systems that can be used by clinicians to analyse ulcer condition. Automated analysis and recognition of DFU can help to improve the diagnosis of DFUs. The next challenge (DFUC2021 \cite{yap2020dfuc2021}) will focus on multi-class DFU recognition.
	\item  With the growth in the number of people diagnosed with diabetes, remote detection and monitoring of DFU can reduce the burden on health services. Research in optimization of deep learning models for remote monitoring is another active research area that has the potential to change the healthcare landscape globally.
\end{itemize}

Due to accompanying and challenging phenomena, such as malformed toenails, rhagades and hyperkeratosis, the wound area needs to be distinguished sufficiently first. Without an accurate deep learning algorithm for DFU detection, a reliable and performant segmentation and accurate wound size estimation is not possible. A reliable detection on typical wound care documentation images, created under uncontrolled (non-laboratory) conditions, remains the first and cardinal problem.

\section*{Acknowledgments}
We gratefully acknowledge the support of NVIDIA Corporation for the use of GPUs for this challenge and sponsoring our event. A.A., D.B.A. and M.O. were supported by the National Health and Medical Research Council [GNT1174405] and the Victorian Government's OIS Program.

\bibliographystyle{model2-names.bst}\biboptions{authoryear}
\bibliography{Ref}

\begin{thebibliography}{57}
\expandafter\ifx\csname natexlab\endcsname\relax\def\natexlab#1{#1}\fi
\providecommand{\url}[1]{\texttt{#1}}
\providecommand{\href}[2]{#2}
\providecommand{\path}[1]{#1}
\providecommand{\DOIprefix}{doi:}
\providecommand{\ArXivprefix}{arXiv:}
\providecommand{\URLprefix}{URL: }
\providecommand{\Pubmedprefix}{pmid:}
\providecommand{\doi}[1]{\href{http://dx.doi.org/#1}{\path{#1}}}
\providecommand{\Pubmed}[1]{\href{pmid:#1}{\path{#1}}}
\providecommand{\bibinfo}[2]{#2}
\ifx\xfnm\relax \def\xfnm[#1]{\unskip,\space#1}\fi
\bibitem[{Armstrong et~al.(2017)Armstrong, Boulton and
  Bus}]{armstrong2017diabetic}
\bibinfo{author}{Armstrong, D.G.}, \bibinfo{author}{Boulton, A.J.},
  \bibinfo{author}{Bus, S.A.}, \bibinfo{year}{2017}.
\newblock \bibinfo{title}{{Diabetic Foot Ulcers and Their Recurrence}}.
\newblock \bibinfo{journal}{New England Journal of Medicine}
  \bibinfo{volume}{376}, \bibinfo{pages}{2367--2375}.
\newblock \DOIprefix\doi{10.1056/nejmra1615439}.
\bibitem[{Bochkovskiy et~al.(2020)Bochkovskiy, Wang and Liao}]{yolov4}
\bibinfo{author}{Bochkovskiy, A.}, \bibinfo{author}{Wang, C.Y.},
  \bibinfo{author}{Liao, H.Y.M.}, \bibinfo{year}{2020}.
\newblock \bibinfo{title}{{YOLOv4: Optimal Speed and Accuracy of Object
  Detection}}.
\newblock \href{http://arxiv.org/abs/2004.10934}{\tt arXiv:2004.10934}.
\bibitem[{Bodla et~al.(2017)Bodla, Singh, Chellappa and Davis}]{bodla2017soft}
\bibinfo{author}{Bodla, N.}, \bibinfo{author}{Singh, B.},
  \bibinfo{author}{Chellappa, R.}, \bibinfo{author}{Davis, L.S.},
  \bibinfo{year}{2017}.
\newblock \bibinfo{title}{{Soft-{NMS} {\textemdash} Improving Object Detection
  with One Line of Code}}, in: \bibinfo{booktitle}{2017 {IEEE} International
  Conference on Computer Vision ({ICCV})}, \bibinfo{publisher}{{IEEE}}. pp.
  \bibinfo{pages}{5561--5569}.
\newblock \DOIprefix\doi{10.1109/iccv.2017.593}.
\bibitem[{Brown et~al.(2017)Brown, Ploderer, Da~Seng, Lazzarini and van
  Netten}]{brown2017myfootcare}
\bibinfo{author}{Brown, R.}, \bibinfo{author}{Ploderer, B.},
  \bibinfo{author}{Da~Seng, L.S.}, \bibinfo{author}{Lazzarini, P.},
  \bibinfo{author}{van Netten, J.}, \bibinfo{year}{2017}.
\newblock \bibinfo{title}{{MyFootCare: A Mobile Self-Tracking Tool to Promote
  Self-Care amongst People with Diabetic Foot Ulcers}}, in:
  \bibinfo{booktitle}{Proceedings of the 29th Australian Conference on
  Computer-Human Interaction (OzCHI'17)}, \bibinfo{publisher}{Association for
  Computing Machinery}. p. \bibinfo{pages}{462–466}.
\newblock \DOIprefix\doi{10.1145/3152771.3156158}.
\bibitem[{Br\"ungel and Friedrich(2021)}]{bruengel2021detryolov5}
\bibinfo{author}{Br\"ungel, R.}, \bibinfo{author}{Friedrich, C.M.},
  \bibinfo{year}{2021}.
\newblock \bibinfo{title}{Detr and yolov5: Exploring performance and
  self-training for diabetic foot ulcer detection}, in:
  \bibinfo{booktitle}{2021 IEEE 34rd International Symposium on Computer-Based
  Medical Systems (CBMS)}, \bibinfo{publisher}{{IEEE}}.
\newblock \bibinfo{note}{Accepted for presentation.}
\bibitem[{Buades et~al.(2005)Buades, Coll and Morel}]{buades2005}
\bibinfo{author}{Buades, A.}, \bibinfo{author}{Coll, B.},
  \bibinfo{author}{Morel, J.M.}, \bibinfo{year}{2005}.
\newblock \bibinfo{title}{{A Non-Local Algorithm for Image Denoising}}, in:
  \bibinfo{booktitle}{2005 {IEEE} Computer Society Conference on Computer
  Vision and Pattern Recognition ({CVPR}'05)}, \bibinfo{publisher}{{IEEE}}. pp.
  \bibinfo{pages}{60--65}.
\newblock \DOIprefix\doi{10.1109/cvpr.2005.38}.
\bibitem[{Cai and Vasconcelos(2018)}]{Cai2017Cascade}
\bibinfo{author}{Cai, Z.}, \bibinfo{author}{Vasconcelos, N.},
  \bibinfo{year}{2018}.
\newblock \bibinfo{title}{{Cascade R-CNN: Delving Into High Quality Object
  Detection}}, in: \bibinfo{booktitle}{2018 {IEEE}/{CVF} Conference on Computer
  Vision and Pattern Recognition}, \bibinfo{publisher}{{IEEE}}. pp.
  \bibinfo{pages}{6154--6162}.
\newblock \DOIprefix\doi{10.1109/cvpr.2018.00644}.
\bibitem[{Cai and Vasconcelos(2021)}]{cai2019cascade}
\bibinfo{author}{Cai, Z.}, \bibinfo{author}{Vasconcelos, N.},
  \bibinfo{year}{2021}.
\newblock \bibinfo{title}{{Cascade R-{CNN}: High Quality Object Detection and
  Instance Segmentation}}.
\newblock \bibinfo{journal}{{IEEE} Transactions on Pattern Analysis and Machine
  Intelligence} \bibinfo{volume}{43}, \bibinfo{pages}{1483--1498}.
\newblock \DOIprefix\doi{10.1109/tpami.2019.2956516}.
\bibitem[{Cao et~al.(2020)Cao, Chen, Loy and Lin}]{Cao2020PISA}
\bibinfo{author}{Cao, Y.}, \bibinfo{author}{Chen, K.}, \bibinfo{author}{Loy,
  C.C.}, \bibinfo{author}{Lin, D.}, \bibinfo{year}{2020}.
\newblock \bibinfo{title}{Prime sample attention in object detection}, in:
  \bibinfo{booktitle}{2020 {IEEE}/{CVF} Conference on Computer Vision and
  Pattern Recognition ({CVPR})}, \bibinfo{publisher}{{IEEE}}. pp.
  \bibinfo{pages}{11580--11588}.
\newblock \DOIprefix\doi{10.1109/cvpr42600.2020.01160}.
\bibitem[{Cassidy et~al.(2020)Cassidy, Reeves, Joseph, Gillespie, O'Shea,
  Rajbhandari, Maiya, Frank, Boulton, Armstrong, Najafi, Wu and
  Yap}]{cassidy2020dfuc2020}
\bibinfo{author}{Cassidy, B.}, \bibinfo{author}{Reeves, N.D.},
  \bibinfo{author}{Joseph, P.}, \bibinfo{author}{Gillespie, D.},
  \bibinfo{author}{O'Shea, C.}, \bibinfo{author}{Rajbhandari, S.},
  \bibinfo{author}{Maiya, A.G.}, \bibinfo{author}{Frank, E.},
  \bibinfo{author}{Boulton, A.}, \bibinfo{author}{Armstrong, D.},
  \bibinfo{author}{Najafi, B.}, \bibinfo{author}{Wu, J.}, \bibinfo{author}{Yap,
  M.H.}, \bibinfo{year}{2020}.
\newblock \bibinfo{title}{{DFUC2020: Analysis Towards Diabetic Foot Ulcer
  Detection}} \href{http://arxiv.org/abs/2004.11853}{\tt arXiv:2004.11853}.
\bibitem[{Drukker et~al.(2002)Drukker, Giger, Horsch, Kupinski, Vyborny and
  Mendelson}]{drukker2002computerized}
\bibinfo{author}{Drukker, K.}, \bibinfo{author}{Giger, M.L.},
  \bibinfo{author}{Horsch, K.}, \bibinfo{author}{Kupinski, M.A.},
  \bibinfo{author}{Vyborny, C.J.}, \bibinfo{author}{Mendelson, E.B.},
  \bibinfo{year}{2002}.
\newblock \bibinfo{title}{Computerized lesion detection on breast ultrasound}.
\newblock \bibinfo{journal}{Medical Physics} \bibinfo{volume}{29},
  \bibinfo{pages}{1438--1446}.
\newblock \DOIprefix\doi{10.1118/1.1485995}.
\bibitem[{Goyal and Hassanpour(2020)}]{goyal2020refined}
\bibinfo{author}{Goyal, M.}, \bibinfo{author}{Hassanpour, S.},
  \bibinfo{year}{2020}.
\newblock \bibinfo{title}{{A Refined Deep Learning Architecture for Diabetic
  Foot Ulcers Detection}} \href{http://arxiv.org/abs/2007.07922}{\tt
  arXiv:2007.07922}.
\bibitem[{Goyal et~al.(2019a)Goyal, Hassanpour and Yap}]{goyal2018region}
\bibinfo{author}{Goyal, M.}, \bibinfo{author}{Hassanpour, S.},
  \bibinfo{author}{Yap, M.H.}, \bibinfo{year}{2019}a.
\newblock \bibinfo{title}{{Region of Interest Detection in Dermoscopic Images
  for Natural Data-augmentation}} \href{http://arxiv.org/abs/1807.10711}{\tt
  arXiv:1807.10711}.
\bibitem[{Goyal et~al.(2020a)Goyal, Reeves, Davison, Rajbhandari, Spragg and
  Yap}]{goyal2018dfunet}
\bibinfo{author}{Goyal, M.}, \bibinfo{author}{Reeves, N.D.},
  \bibinfo{author}{Davison, A.K.}, \bibinfo{author}{Rajbhandari, S.},
  \bibinfo{author}{Spragg, J.}, \bibinfo{author}{Yap, M.H.},
  \bibinfo{year}{2020}a.
\newblock \bibinfo{title}{{DFUNet: Convolutional Neural Networks for Diabetic
  Foot Ulcer Classification}}.
\newblock \bibinfo{journal}{IEEE Transactions on Emerging Topics in
  Computational Intelligence} \bibinfo{volume}{4}, \bibinfo{pages}{728--739}.
\newblock \DOIprefix\doi{10.1109/TETCI.2018.2866254}.
\bibitem[{Goyal et~al.(2020b)Goyal, Reeves, Rajbhandari, Ahmad, Wang and
  Yap}]{goyal2020recognition}
\bibinfo{author}{Goyal, M.}, \bibinfo{author}{Reeves, N.D.},
  \bibinfo{author}{Rajbhandari, S.}, \bibinfo{author}{Ahmad, N.},
  \bibinfo{author}{Wang, C.}, \bibinfo{author}{Yap, M.H.},
  \bibinfo{year}{2020}b.
\newblock \bibinfo{title}{Recognition of ischaemia and infection in diabetic
  foot ulcers: Dataset and techniques}.
\newblock \bibinfo{journal}{Computers in Biology and Medicine}
  \bibinfo{volume}{117}, \bibinfo{pages}{103616}.
\newblock \DOIprefix\doi{10.1016/j.compbiomed.2020.103616}.
\bibitem[{Goyal et~al.(2019b)Goyal, Reeves, Rajbhandari and
  Yap}]{goyal2018robust}
\bibinfo{author}{Goyal, M.}, \bibinfo{author}{Reeves, N.D.},
  \bibinfo{author}{Rajbhandari, S.}, \bibinfo{author}{Yap, M.H.},
  \bibinfo{year}{2019}b.
\newblock \bibinfo{title}{{Robust Methods for Real-Time Diabetic Foot Ulcer
  Detection and Localization on Mobile Devices}}.
\newblock \bibinfo{journal}{IEEE Journal of Biomedical and Health Informatics}
  \bibinfo{volume}{23}, \bibinfo{pages}{1730--1741}.
\newblock \DOIprefix\doi{10.1109/JBHI.2018.2868656}.
\bibitem[{Goyal et~al.(2017)Goyal, Yap, Reeves, Rajbhandari and
  Spragg}]{goyal2017fully}
\bibinfo{author}{Goyal, M.}, \bibinfo{author}{Yap, M.H.},
  \bibinfo{author}{Reeves, N.D.}, \bibinfo{author}{Rajbhandari, S.},
  \bibinfo{author}{Spragg, J.}, \bibinfo{year}{2017}.
\newblock \bibinfo{title}{Fully convolutional networks for diabetic foot ulcer
  segmentation}, in: \bibinfo{booktitle}{2017 IEEE International Conference on
  Systems, Man, and Cybernetics (SMC)}, \bibinfo{publisher}{{IEEE}}. pp.
  \bibinfo{pages}{618--623}.
\newblock \DOIprefix\doi{10.1109/SMC.2017.8122675}.
\bibitem[{{He} et~al.(2017){He}, {Gkioxari}, {Dollár} and
  {Girshick}}]{MaskRCNN}
\bibinfo{author}{{He}, K.}, \bibinfo{author}{{Gkioxari}, G.},
  \bibinfo{author}{{Dollár}, P.}, \bibinfo{author}{{Girshick}, R.},
  \bibinfo{year}{2017}.
\newblock \bibinfo{title}{{Mask R-CNN}}, in: \bibinfo{booktitle}{2017 IEEE
  International Conference on Computer Vision (ICCV)},
  \bibinfo{publisher}{{IEEE}}. pp. \bibinfo{pages}{2980--2988}.
\newblock \DOIprefix\doi{10.1109/ICCV.2017.322}.
\bibitem[{He et~al.(2015)He, Zhang, Ren and Sun}]{he2015spatial}
\bibinfo{author}{He, K.}, \bibinfo{author}{Zhang, X.}, \bibinfo{author}{Ren,
  S.}, \bibinfo{author}{Sun, J.}, \bibinfo{year}{2015}.
\newblock \bibinfo{title}{{Spatial Pyramid Pooling in Deep Convolutional
  Networks for Visual Recognition}}.
\newblock \bibinfo{journal}{{IEEE} Transactions on Pattern Analysis and Machine
  Intelligence} \bibinfo{volume}{37}, \bibinfo{pages}{1904--1916}.
\newblock \DOIprefix\doi{10.1109/tpami.2015.2389824}.
\bibitem[{Hendrycks and Gimpel(2020)}]{hendrycks2020gaussian}
\bibinfo{author}{Hendrycks, D.}, \bibinfo{author}{Gimpel, K.},
  \bibinfo{year}{2020}.
\newblock \bibinfo{title}{{Gaussian Error Linear Units (GELUs)}}.
\newblock \bibinfo{journal}{arXiv:1606.08415}
  \href{http://arxiv.org/abs/1606.08415}{\tt arXiv:1606.08415}.
\bibitem[{Jocher et~al.(2020a)Jocher, Changyu, Hogan, Yu, changyu98, Rai and
  Sullivan}]{jocher2020a}
\bibinfo{author}{Jocher, G.}, \bibinfo{author}{Changyu, L.},
  \bibinfo{author}{Hogan, A.}, \bibinfo{author}{Yu, L.},
  \bibinfo{author}{changyu98}, \bibinfo{author}{Rai, P.},
  \bibinfo{author}{Sullivan, T.}, \bibinfo{year}{2020}a.
\newblock \bibinfo{title}{ultralytics/yolov5: Initial release}.
\newblock \DOIprefix\doi{10.5281/zenodo.3908560}.
\bibitem[{Jocher et~al.(2020b)Jocher, Kwon, guigarfr, Veitch-Michaelis,
  perry0418, Ttayu, Marc, Bianconi, Baltacı, Suess, Chen, Yang, idow09,
  WannaSeaU, Xinyu, Shead, Havlik, Skalski, NirZarrabi, LukeAI, LinCoce, Hu,
  IlyaOvodov, GoogleWiki, Reveriano, Falak and Kendall}]{jocher2020b}
\bibinfo{author}{Jocher, G.}, \bibinfo{author}{Kwon, Y.},
  \bibinfo{author}{guigarfr}, \bibinfo{author}{Veitch-Michaelis, J.},
  \bibinfo{author}{perry0418}, \bibinfo{author}{Ttayu}, \bibinfo{author}{Marc},
  \bibinfo{author}{Bianconi, G.}, \bibinfo{author}{Baltacı, F.},
  \bibinfo{author}{Suess, D.}, \bibinfo{author}{Chen, T.},
  \bibinfo{author}{Yang, P.}, \bibinfo{author}{idow09},
  \bibinfo{author}{WannaSeaU}, \bibinfo{author}{Xinyu, W.},
  \bibinfo{author}{Shead, T.M.}, \bibinfo{author}{Havlik, T.},
  \bibinfo{author}{Skalski, P.}, \bibinfo{author}{NirZarrabi},
  \bibinfo{author}{LukeAI}, \bibinfo{author}{LinCoce}, \bibinfo{author}{Hu,
  J.}, \bibinfo{author}{IlyaOvodov}, \bibinfo{author}{GoogleWiki},
  \bibinfo{author}{Reveriano, F.}, \bibinfo{author}{Falak},
  \bibinfo{author}{Kendall, D.}, \bibinfo{year}{2020}b.
\newblock \bibinfo{title}{{ultralytics/yolov3: 43.1mAP@0.5:0.95 on COCO2014}}.
\newblock \DOIprefix\doi{10.5281/zenodo.3785397}.
\bibitem[{Jocher et~al.(2021a)Jocher, Stoken, Borovec, NanoCode012, Chaurasia,
  TaoXie, Changyu, V, Laughing, tkianai, yxNONG, Hogan, lorenzomammana,
  AlexWang1900, Hajek, Diaconu, Marc, Kwon, oleg, wanghaoyang0106, Defretin,
  Lohia, ml5ah, Milanko, Fineran, Khromov, Yiwei, Doug, Durgesh and
  Ingham}]{jocher2021}
\bibinfo{author}{Jocher, G.}, \bibinfo{author}{Stoken, A.},
  \bibinfo{author}{Borovec, J.}, \bibinfo{author}{NanoCode012},
  \bibinfo{author}{Chaurasia, A.}, \bibinfo{author}{TaoXie},
  \bibinfo{author}{Changyu, L.}, \bibinfo{author}{V, A.},
  \bibinfo{author}{Laughing}, \bibinfo{author}{tkianai},
  \bibinfo{author}{yxNONG}, \bibinfo{author}{Hogan, A.},
  \bibinfo{author}{lorenzomammana}, \bibinfo{author}{AlexWang1900},
  \bibinfo{author}{Hajek, J.}, \bibinfo{author}{Diaconu, L.},
  \bibinfo{author}{Marc}, \bibinfo{author}{Kwon, Y.}, \bibinfo{author}{oleg},
  \bibinfo{author}{wanghaoyang0106}, \bibinfo{author}{Defretin, Y.},
  \bibinfo{author}{Lohia, A.}, \bibinfo{author}{ml5ah},
  \bibinfo{author}{Milanko, B.}, \bibinfo{author}{Fineran, B.},
  \bibinfo{author}{Khromov, D.}, \bibinfo{author}{Yiwei, D.},
  \bibinfo{author}{Doug}, \bibinfo{author}{Durgesh}, \bibinfo{author}{Ingham,
  F.}, \bibinfo{year}{2021}a.
\newblock \bibinfo{title}{{ultralytics/yolov5: v5.0 - YOLOv5-P6 1280 models,
  AWS, Supervise.ly and YouTube integrations}}.
\newblock \DOIprefix\doi{10.5281/zenodo.4679653}.
\bibitem[{Jocher et~al.(2021b)Jocher, Stoken, Borovec, NanoCode012,
  ChristopherSTAN, Changyu, Laughing, tkianai, yxNONG, Hogan, lorenzomammana,
  AlexWang1900, Chaurasia, Diaconu, Marc, wanghaoyang0106, ml5ah, Doug,
  Durgesh, Ingham, Frederik, Guilhen, Colmagro, Ye, Jacobsolawetz, Poznanski,
  Fang, Kim, Doan and Yu}]{glenn_jocher_2021_4418161}
\bibinfo{author}{Jocher, G.}, \bibinfo{author}{Stoken, A.},
  \bibinfo{author}{Borovec, J.}, \bibinfo{author}{NanoCode012},
  \bibinfo{author}{ChristopherSTAN}, \bibinfo{author}{Changyu, L.},
  \bibinfo{author}{Laughing}, \bibinfo{author}{tkianai},
  \bibinfo{author}{yxNONG}, \bibinfo{author}{Hogan, A.},
  \bibinfo{author}{lorenzomammana}, \bibinfo{author}{AlexWang1900},
  \bibinfo{author}{Chaurasia, A.}, \bibinfo{author}{Diaconu, L.},
  \bibinfo{author}{Marc}, \bibinfo{author}{wanghaoyang0106},
  \bibinfo{author}{ml5ah}, \bibinfo{author}{Doug}, \bibinfo{author}{Durgesh},
  \bibinfo{author}{Ingham, F.}, \bibinfo{author}{Frederik},
  \bibinfo{author}{Guilhen}, \bibinfo{author}{Colmagro, A.},
  \bibinfo{author}{Ye, H.}, \bibinfo{author}{Jacobsolawetz},
  \bibinfo{author}{Poznanski, J.}, \bibinfo{author}{Fang, J.},
  \bibinfo{author}{Kim, J.}, \bibinfo{author}{Doan, K.}, \bibinfo{author}{Yu,
  L.}, \bibinfo{year}{2021}b.
\newblock \bibinfo{title}{{ultralytics/yolov5: v4.0 - nn.SiLU() activations,
  Weights \& Biases logging, PyTorch Hub integration}}.
\newblock \DOIprefix\doi{10.5281/zenodo.4418161}.
\bibitem[{Koitka and Friedrich(2017)}]{koitka2017}
\bibinfo{author}{Koitka, S.}, \bibinfo{author}{Friedrich, C.M.},
  \bibinfo{year}{2017}.
\newblock \bibinfo{title}{{Optimized Convolutional Neural Network Ensembles for
  Medical Subfigure Classification}}, in: \bibinfo{booktitle}{In Experimental
  IR Meets Multilinguality, Multimodality, and Interaction 8th International
  Conference of the CLEF Association, CLEF 2017, Lecture Notes in Computer
  Science (LNCS)}. \bibinfo{publisher}{Springer International Publishing}, pp.
  \bibinfo{pages}{57--68}.
\newblock \DOIprefix\doi{10.1007/978-3-319-65813-1\_5}.
\bibitem[{Li et~al.(2018)Li, Peng, Yu, Zhang, Deng and Sun}]{2018DetNet}
\bibinfo{author}{Li, Z.}, \bibinfo{author}{Peng, C.}, \bibinfo{author}{Yu, G.},
  \bibinfo{author}{Zhang, X.}, \bibinfo{author}{Deng, Y.},
  \bibinfo{author}{Sun, J.}, \bibinfo{year}{2018}.
\newblock \bibinfo{title}{{{DetNet}: Design Backbone for Object Detection}},
  in: \bibinfo{booktitle}{European Conference on Computer Vision ({ECCV})
  2018}. \bibinfo{publisher}{Springer International Publishing}, pp.
  \bibinfo{pages}{339--354}.
\newblock \DOIprefix\doi{10.1007/978-3-030-01240-3_21}.
\bibitem[{Lin et~al.(2017)Lin, Dollár, Girshick, He, Hariharan and
  Belongie}]{aa_lin2017feature}
\bibinfo{author}{Lin, T.Y.}, \bibinfo{author}{Dollár, P.},
  \bibinfo{author}{Girshick, R.}, \bibinfo{author}{He, K.},
  \bibinfo{author}{Hariharan, B.}, \bibinfo{author}{Belongie, S.},
  \bibinfo{year}{2017}.
\newblock \bibinfo{title}{{Feature Pyramid Networks for Object Detection}}, in:
  \bibinfo{booktitle}{2017 IEEE Conference on Computer Vision and Pattern
  Recognition (CVPR)}, \bibinfo{publisher}{{IEEE}}. pp.
  \bibinfo{pages}{936--944}.
\newblock \DOIprefix\doi{10.1109/CVPR.2017.106}.
\bibitem[{Lin et~al.(2014)Lin, Maire, Belongie, Hays, Perona, Ramanan,
  Doll{\'{a}}r and Zitnick}]{lin2014microsoft}
\bibinfo{author}{Lin, T.Y.}, \bibinfo{author}{Maire, M.},
  \bibinfo{author}{Belongie, S.}, \bibinfo{author}{Hays, J.},
  \bibinfo{author}{Perona, P.}, \bibinfo{author}{Ramanan, D.},
  \bibinfo{author}{Doll{\'{a}}r, P.}, \bibinfo{author}{Zitnick, C.L.},
  \bibinfo{year}{2014}.
\newblock \bibinfo{title}{{Microsoft {COCO}: Common Objects in Context}}, in:
  \bibinfo{booktitle}{European Conference on Computer Vision ({ECCV}) 2014},
  \bibinfo{publisher}{Springer International Publishing}. pp.
  \bibinfo{pages}{740--755}.
\newblock \DOIprefix\doi{10.1007/978-3-319-10602-1_48}.
\bibitem[{Liu et~al.(2018)Liu, Qi, Qin, Shi and Jia}]{liu2018path}
\bibinfo{author}{Liu, S.}, \bibinfo{author}{Qi, L.}, \bibinfo{author}{Qin, H.},
  \bibinfo{author}{Shi, J.}, \bibinfo{author}{Jia, J.}, \bibinfo{year}{2018}.
\newblock \bibinfo{title}{{Path Aggregation Network for Instance
  Segmentation}}, in: \bibinfo{booktitle}{2018 {IEEE}/{CVF} Conference on
  Computer Vision and Pattern Recognition}, \bibinfo{publisher}{{IEEE}}. pp.
  \bibinfo{pages}{8759--8768}.
\newblock \DOIprefix\doi{10.1109/cvpr.2018.00913}.
\bibitem[{Maas et~al.(2013)Maas, Hannun and Ng}]{maas2013leaky_relu}
\bibinfo{author}{Maas, A.L.}, \bibinfo{author}{Hannun, A.Y.},
  \bibinfo{author}{Ng, A.Y.}, \bibinfo{year}{2013}.
\newblock \bibinfo{title}{{Rectifier Nonlinearities Improve Neural Network
  Acoustic Models}}, in: \bibinfo{booktitle}{{Proceedings of the 30th
  International Conference on Machine Learning (ICML) 2013}},
  \bibinfo{publisher}{ICML}.
\bibitem[{Ng et~al.(2019)Ng, Goyal, Hewitt and Yap}]{hua2019effect}
\bibinfo{author}{Ng, J.H.}, \bibinfo{author}{Goyal, M.},
  \bibinfo{author}{Hewitt, B.}, \bibinfo{author}{Yap, M.H.},
  \bibinfo{year}{2019}.
\newblock \bibinfo{title}{{The effect of color constancy algorithms on semantic
  segmentation of skin lesions}}, in: \bibinfo{booktitle}{Medical Imaging 2019:
  Biomedical Applications in Molecular, Structural, and Functional Imaging},
  \bibinfo{organization}{International Society for Optics and Photonics}.
  \bibinfo{publisher}{SPIE}. pp. \bibinfo{pages}{138--145}.
\newblock \DOIprefix\doi{10.1117/12.2512702}.
\bibitem[{Paszke et~al.(2019)Paszke, Gross, Massa, Lerer, Bradbury, Chanan,
  Killeen, Lin, Gimelshein, Antiga, Desmaison, Kopf, Yang, DeVito, Raison,
  Tejani, Chilamkurthy, Steiner, Fang, Bai and Chintala}]{pytorch}
\bibinfo{author}{Paszke, A.}, \bibinfo{author}{Gross, S.},
  \bibinfo{author}{Massa, F.}, \bibinfo{author}{Lerer, A.},
  \bibinfo{author}{Bradbury, J.}, \bibinfo{author}{Chanan, G.},
  \bibinfo{author}{Killeen, T.}, \bibinfo{author}{Lin, Z.},
  \bibinfo{author}{Gimelshein, N.}, \bibinfo{author}{Antiga, L.},
  \bibinfo{author}{Desmaison, A.}, \bibinfo{author}{Kopf, A.},
  \bibinfo{author}{Yang, E.}, \bibinfo{author}{DeVito, Z.},
  \bibinfo{author}{Raison, M.}, \bibinfo{author}{Tejani, A.},
  \bibinfo{author}{Chilamkurthy, S.}, \bibinfo{author}{Steiner, B.},
  \bibinfo{author}{Fang, L.}, \bibinfo{author}{Bai, J.},
  \bibinfo{author}{Chintala, S.}, \bibinfo{year}{2019}.
\newblock \bibinfo{title}{Pytorch: An imperative style, high-performance deep
  learning library}, in: \bibinfo{booktitle}{Advances in Neural Information
  Processing Systems 32 (NeurIPS 2019)}, \bibinfo{publisher}{Curran Associates,
  Inc.}. pp. \bibinfo{pages}{8024--8035}.
\bibitem[{Pebesma(2018)}]{Rsf}
\bibinfo{author}{Pebesma, E.}, \bibinfo{year}{2018}.
\newblock \bibinfo{title}{{Simple Features for R: Standardized Support for
  Spatial Vector Data}}.
\newblock \bibinfo{journal}{The R Journal} \bibinfo{volume}{10},
  \bibinfo{pages}{439--446}.
\newblock \DOIprefix\doi{10.32614/RJ-2018-009}.
\bibitem[{{R Core Team}(2020)}]{Rlang}
\bibinfo{author}{{R Core Team}}, \bibinfo{year}{2020}.
\newblock \bibinfo{title}{{R: A Language and Environment for Statistical
  Computing}}.
\newblock \bibinfo{organization}{R Foundation for Statistical Computing}.
  \bibinfo{address}{Vienna, Austria}.
\newblock \URLprefix \url{https://www.R-project.org/}.
  \bibinfo{note}{{Retrieved on 2021-04-28}}.
\bibitem[{Redmon et~al.(2016)Redmon, Divvala, Girshick and Farhadi}]{yolov1}
\bibinfo{author}{Redmon, J.}, \bibinfo{author}{Divvala, S.},
  \bibinfo{author}{Girshick, R.}, \bibinfo{author}{Farhadi, A.},
  \bibinfo{year}{2016}.
\newblock \bibinfo{title}{{You Only Look Once: Unified, Real-Time Object
  Detection}}, in: \bibinfo{booktitle}{2016 {IEEE} Conference on Computer
  Vision and Pattern Recognition ({CVPR})}, \bibinfo{publisher}{{IEEE}}. pp.
  \bibinfo{pages}{779--788}.
\newblock \DOIprefix\doi{10.1109/cvpr.2016.91}.
\bibitem[{Redmon and Farhadi(2017)}]{yolov2}
\bibinfo{author}{Redmon, J.}, \bibinfo{author}{Farhadi, A.},
  \bibinfo{year}{2017}.
\newblock \bibinfo{title}{{{YOLO}9000: Better, Faster, Stronger}}, in:
  \bibinfo{booktitle}{2017 {IEEE} Conference on Computer Vision and Pattern
  Recognition ({CVPR})}, \bibinfo{publisher}{{IEEE}}. pp.
  \bibinfo{pages}{7263--7271}.
\newblock \DOIprefix\doi{10.1109/cvpr.2017.690}.
\bibitem[{Redmon and Farhadi(2018)}]{aa_redmon2018yolov3}
\bibinfo{author}{Redmon, J.}, \bibinfo{author}{Farhadi, A.},
  \bibinfo{year}{2018}.
\newblock \bibinfo{title}{{YOLOv3: An Incremental Improvement}}
  \href{http://arxiv.org/abs/1804.02767}{\tt arXiv:1804.02767}.
\bibitem[{Ren et~al.(2017)Ren, He, Girshick and Sun}]{ren2015faster}
\bibinfo{author}{Ren, S.}, \bibinfo{author}{He, K.}, \bibinfo{author}{Girshick,
  R.}, \bibinfo{author}{Sun, J.}, \bibinfo{year}{2017}.
\newblock \bibinfo{title}{{Faster R-{CNN}: Towards Real-Time Object Detection
  with Region Proposal Networks}}.
\newblock \bibinfo{journal}{{IEEE} Transactions on Pattern Analysis and Machine
  Intelligence} \bibinfo{volume}{39}, \bibinfo{pages}{1137--1149}.
\newblock \DOIprefix\doi{10.1109/tpami.2016.2577031}.
\bibitem[{Saeedi et~al.(2019)Saeedi, Petersohn, Salpea, Malanda, Karuranga,
  Unwin, Colagiuri, Guariguata, Motala, Ogurtsova, Shaw, Bright and
  Williams}]{IDF}
\bibinfo{author}{Saeedi, P.}, \bibinfo{author}{Petersohn, I.},
  \bibinfo{author}{Salpea, P.}, \bibinfo{author}{Malanda, B.},
  \bibinfo{author}{Karuranga, S.}, \bibinfo{author}{Unwin, N.},
  \bibinfo{author}{Colagiuri, S.}, \bibinfo{author}{Guariguata, L.},
  \bibinfo{author}{Motala, A.A.}, \bibinfo{author}{Ogurtsova, K.},
  \bibinfo{author}{Shaw, J.E.}, \bibinfo{author}{Bright, D.},
  \bibinfo{author}{Williams, R.}, \bibinfo{year}{2019}.
\newblock \bibinfo{title}{{Global and regional diabetes prevalence estimates
  for 2019 and projections for 2030 and 2045: Results from the International
  Diabetes Federation Diabetes Atlas, 9th edition}}.
\newblock \bibinfo{journal}{Diabetes Research and Clinical Practice}
  \bibinfo{volume}{157}, \bibinfo{pages}{107843}.
\newblock \DOIprefix\doi{10.1016/j.diabres.2019.107843}.
\bibitem[{Solovyev et~al.(2021)Solovyev, Wang and
  Gabruseva}]{solovyev2019weighted}
\bibinfo{author}{Solovyev, R.}, \bibinfo{author}{Wang, W.},
  \bibinfo{author}{Gabruseva, T.}, \bibinfo{year}{2021}.
\newblock \bibinfo{title}{Weighted boxes fusion: Ensembling boxes from
  different object detection models}.
\newblock \bibinfo{journal}{Image and Vision Computing} \bibinfo{volume}{107},
  \bibinfo{pages}{104117}.
\newblock \DOIprefix\doi{10.1016/j.imavis.2021.104117}.
\bibitem[{Tan and Le(2019)}]{tan2019effnet}
\bibinfo{author}{Tan, M.}, \bibinfo{author}{Le, Q.}, \bibinfo{year}{2019}.
\newblock \bibinfo{title}{{{E}fficient{N}et: Rethinking Model Scaling for
  Convolutional Neural Networks}}, in: \bibinfo{booktitle}{Proceedings of the
  36th International Conference on Machine Learning},
  \bibinfo{publisher}{PMLR}. pp. \bibinfo{pages}{6105--6114}.
\newblock \URLprefix \url{http://proceedings.mlr.press/v97/tan19a.html}.
  \bibinfo{note}{{Retrieved on 2021-04-28}}.
\bibitem[{Tan et~al.(2020)Tan, Pang and Le}]{tan2020efficientdet}
\bibinfo{author}{Tan, M.}, \bibinfo{author}{Pang, R.}, \bibinfo{author}{Le,
  Q.V.}, \bibinfo{year}{2020}.
\newblock \bibinfo{title}{{EfficientDet: Scalable and Efficient Object
  Detection}}, in: \bibinfo{booktitle}{2020 IEEE/CVF Conference on Computer
  Vision and Pattern Recognition (CVPR)}, \bibinfo{publisher}{{IEEE}}. pp.
  \bibinfo{pages}{10778--10787}.
\newblock \DOIprefix\doi{10.1109/CVPR42600.2020.01079}.
\bibitem[{Wang et~al.(2020)Wang, Mark~Liao, Wu, Chen, Hsieh and
  Yeh}]{wang2020cspnet}
\bibinfo{author}{Wang, C.Y.}, \bibinfo{author}{Mark~Liao, H.Y.},
  \bibinfo{author}{Wu, Y.H.}, \bibinfo{author}{Chen, P.Y.},
  \bibinfo{author}{Hsieh, J.W.}, \bibinfo{author}{Yeh, I.H.},
  \bibinfo{year}{2020}.
\newblock \bibinfo{title}{{CSPNet: A New Backbone that can Enhance Learning
  Capability of CNN}}, in: \bibinfo{booktitle}{2020 IEEE/CVF Conference on
  Computer Vision and Pattern Recognition Workshops (CVPRW)},
  \bibinfo{publisher}{{IEEE}}. pp. \bibinfo{pages}{1571--1580}.
\newblock \DOIprefix\doi{10.1109/CVPRW50498.2020.00203}.
\bibitem[{Wang et~al.(2017a)Wang, Jiang, Qian, Yang, Li, Zhang, Wang and
  Tang}]{2017Residual}
\bibinfo{author}{Wang, F.}, \bibinfo{author}{Jiang, M.}, \bibinfo{author}{Qian,
  C.}, \bibinfo{author}{Yang, S.}, \bibinfo{author}{Li, C.},
  \bibinfo{author}{Zhang, H.}, \bibinfo{author}{Wang, X.},
  \bibinfo{author}{Tang, X.}, \bibinfo{year}{2017}a.
\newblock \bibinfo{title}{{Residual Attention Network for Image
  Classification}}, in: \bibinfo{booktitle}{2017 IEEE Conference on Computer
  Vision and Pattern Recognition (CVPR)}, \bibinfo{publisher}{{IEEE}}. pp.
  \bibinfo{pages}{6450--6458}.
\newblock \DOIprefix\doi{10.1109/CVPR.2017.683}.
\bibitem[{Wang et~al.(2017b)Wang, Pedersen, Agu, Strong and
  Tulu}]{wang2016area}
\bibinfo{author}{Wang, L.}, \bibinfo{author}{Pedersen, P.C.},
  \bibinfo{author}{Agu, E.}, \bibinfo{author}{Strong, D.M.},
  \bibinfo{author}{Tulu, B.}, \bibinfo{year}{2017}b.
\newblock \bibinfo{title}{{Area Determination of Diabetic Foot Ulcer Images
  Using a Cascaded Two-Stage SVM-Based Classification}}.
\newblock \bibinfo{journal}{IEEE Transactions on Biomedical Engineering}
  \bibinfo{volume}{64}, \bibinfo{pages}{2098--2109}.
\newblock \DOIprefix\doi{10.1109/TBME.2016.2632522}.
\bibitem[{Wang et~al.(2015)Wang, Pedersen, Strong, Tulu, Agu and
  Ignotz}]{wang2014smartphone}
\bibinfo{author}{Wang, L.}, \bibinfo{author}{Pedersen, P.C.},
  \bibinfo{author}{Strong, D.M.}, \bibinfo{author}{Tulu, B.},
  \bibinfo{author}{Agu, E.}, \bibinfo{author}{Ignotz, R.},
  \bibinfo{year}{2015}.
\newblock \bibinfo{title}{{Smartphone-Based Wound Assessment System for
  Patients With Diabetes}}.
\newblock \bibinfo{journal}{IEEE Transactions on Biomedical Engineering}
  \bibinfo{volume}{62}, \bibinfo{pages}{477--488}.
\newblock \DOIprefix\doi{10.1109/TBME.2014.2358632}.
\bibitem[{Xie et~al.(2017)Xie, Girshick, Dollár, Tu and
  He}]{Aggregated2017Xie}
\bibinfo{author}{Xie, S.}, \bibinfo{author}{Girshick, R.},
  \bibinfo{author}{Dollár, P.}, \bibinfo{author}{Tu, Z.}, \bibinfo{author}{He,
  K.}, \bibinfo{year}{2017}.
\newblock \bibinfo{title}{{Aggregated Residual Transformations for Deep Neural
  Networks}}, in: \bibinfo{booktitle}{2017 IEEE Conference on Computer Vision
  and Pattern Recognition (CVPR)}, \bibinfo{publisher}{{IEEE}}. pp.
  \bibinfo{pages}{5987--5995}.
\newblock \DOIprefix\doi{10.1109/CVPR.2017.634}.
\bibitem[{Yap et~al.(2008)Yap, Edirisinghe and Bez}]{yap2008novel}
\bibinfo{author}{Yap, M.H.}, \bibinfo{author}{Edirisinghe, E.A.},
  \bibinfo{author}{Bez, H.E.}, \bibinfo{year}{2008}.
\newblock \bibinfo{title}{A novel algorithm for initial lesion detection in
  ultrasound breast images}.
\newblock \bibinfo{journal}{Journal of Applied Clinical Medical Physics}
  \bibinfo{volume}{9}, \bibinfo{pages}{181--199}.
\newblock \DOIprefix\doi{10.1120/jacmp.v9i4.2741}.
\bibitem[{Yap et~al.(2020a)Yap, Goyal, Osman, Mart{\'{\i}}, Denton, Juette and
  Zwiggelaar}]{yap2020breast}
\bibinfo{author}{Yap, M.H.}, \bibinfo{author}{Goyal, M.},
  \bibinfo{author}{Osman, F.}, \bibinfo{author}{Mart{\'{\i}}, R.},
  \bibinfo{author}{Denton, E.}, \bibinfo{author}{Juette, A.},
  \bibinfo{author}{Zwiggelaar, R.}, \bibinfo{year}{2020}a.
\newblock \bibinfo{title}{Breast ultrasound region of interest detection and
  lesion localisation}.
\newblock \bibinfo{journal}{Artificial Intelligence in Medicine}
  \bibinfo{volume}{107}, \bibinfo{pages}{101880}.
\newblock \DOIprefix\doi{10.1016/j.artmed.2020.101880}.
\bibitem[{Yap et~al.(2020b)Yap, Reeves, Boulton, Rajbhandari, Armstrong, Maiya,
  Najafi, Frank and Wu}]{yap2020dfuc2021}
\bibinfo{author}{Yap, M.H.}, \bibinfo{author}{Reeves, N.},
  \bibinfo{author}{Boulton, A.}, \bibinfo{author}{Rajbhandari, S.},
  \bibinfo{author}{Armstrong, D.}, \bibinfo{author}{Maiya, A.G.},
  \bibinfo{author}{Najafi, B.}, \bibinfo{author}{Frank, E.},
  \bibinfo{author}{Wu, J.}, \bibinfo{year}{2020}b.
\newblock \bibinfo{title}{{Diabetic Foot Ulcers Grand Challenge 2021}}.
\newblock \DOIprefix\doi{10.5281/zenodo.3715020}.
\bibitem[{Yap et~al.(2020c)Yap, Reeves, Boulton, Rajbhandari, Armstrong, Maiya,
  Najafi, Frank and Wu}]{yap2020dfuc}
\bibinfo{author}{Yap, M.H.}, \bibinfo{author}{Reeves, N.D.},
  \bibinfo{author}{Boulton, A.}, \bibinfo{author}{Rajbhandari, S.},
  \bibinfo{author}{Armstrong, D.}, \bibinfo{author}{Maiya, A.G.},
  \bibinfo{author}{Najafi, B.}, \bibinfo{author}{Frank, E.},
  \bibinfo{author}{Wu, J.}, \bibinfo{year}{2020}c.
\newblock \bibinfo{title}{{Diabetic Foot Ulcers Grand Challenge 2020}}.
\newblock \DOIprefix\doi{10.5281/zenodo.3715016}.
\bibitem[{Zhang et~al.(2018)Zhang, Cisse, Dauphin and Lopez-Paz}]{mixup}
\bibinfo{author}{Zhang, H.}, \bibinfo{author}{Cisse, M.},
  \bibinfo{author}{Dauphin, Y.N.}, \bibinfo{author}{Lopez-Paz, D.},
  \bibinfo{year}{2018}.
\newblock \bibinfo{title}{{mixup: Beyond Empirical Risk Minimization}}, in:
  \bibinfo{booktitle}{6th International Conference on Learning Representations
  ({ICLR}) 2018}, \bibinfo{publisher}{{ICLR}}.
\newblock \URLprefix \url{https://openreview.net/forum?id=r1Ddp1-Rb}.
  \bibinfo{note}{{Retrieved on 2021-04-28}}.
\bibitem[{Zhang et~al.(2019)Zhang, He, Zhang, Zhang, Xie and Li}]{zhang2019bag}
\bibinfo{author}{Zhang, Z.}, \bibinfo{author}{He, T.}, \bibinfo{author}{Zhang,
  H.}, \bibinfo{author}{Zhang, Z.}, \bibinfo{author}{Xie, J.},
  \bibinfo{author}{Li, M.}, \bibinfo{year}{2019}.
\newblock \bibinfo{title}{{Bag of Freebies for Training Object Detection Neural
  Networks}} \href{http://arxiv.org/abs/1902.04103}{\tt arXiv:1902.04103}.
\bibitem[{Zhao et~al.(2020)Zhao, Huang, Li, Chen and Cheng}]{zhao2020pointer}
\bibinfo{author}{Zhao, W.}, \bibinfo{author}{Huang, H.}, \bibinfo{author}{Li,
  D.}, \bibinfo{author}{Chen, F.}, \bibinfo{author}{Cheng, W.},
  \bibinfo{year}{2020}.
\newblock \bibinfo{title}{{Pointer Defect Detection Based on Transfer Learning
  and Improved Cascade-{RCNN}}}.
\newblock \bibinfo{journal}{Sensors} \bibinfo{volume}{20},
  \bibinfo{pages}{4939}.
\newblock \DOIprefix\doi{10.3390/s20174939}.
\bibitem[{Zhou et~al.(2019)Zhou, Wang and Krähenbühl}]{zhou2019objects}
\bibinfo{author}{Zhou, X.}, \bibinfo{author}{Wang, D.},
  \bibinfo{author}{Krähenbühl, P.}, \bibinfo{year}{2019}.
\newblock \bibinfo{title}{{Objects as Points}}
  \href{http://arxiv.org/abs/1904.07850}{\tt arXiv:1904.07850}.
\bibitem[{Zhu et~al.(2018)Zhu, Fang and Ghamisi}]{zhu2018deformable}
\bibinfo{author}{Zhu, J.}, \bibinfo{author}{Fang, L.},
  \bibinfo{author}{Ghamisi, P.}, \bibinfo{year}{2018}.
\newblock \bibinfo{title}{{Deformable Convolutional Neural Networks for
  Hyperspectral Image Classification}}.
\newblock \bibinfo{journal}{IEEE Geoscience and Remote Sensing Letters}
  \bibinfo{volume}{15}, \bibinfo{pages}{1254--1258}.
\newblock \DOIprefix\doi{10.1109/LGRS.2018.2830403}.
\bibitem[{Zhu et~al.(2019)Zhu, Hu, Lin and Dai}]{Zhu2019Deformable}
\bibinfo{author}{Zhu, X.}, \bibinfo{author}{Hu, H.}, \bibinfo{author}{Lin, S.},
  \bibinfo{author}{Dai, J.}, \bibinfo{year}{2019}.
\newblock \bibinfo{title}{{Deformable ConvNets V2: More Deformable, Better
  Results}}, in: \bibinfo{booktitle}{2019 IEEE/CVF Conference on Computer
  Vision and Pattern Recognition (CVPR)}, \bibinfo{publisher}{{IEEE}}. pp.
  \bibinfo{pages}{9300--9308}.
\newblock \DOIprefix\doi{10.1109/CVPR.2019.00953}.

\end{thebibliography}



\end{document}